\documentclass[lettersize,journal]{IEEEtran}
\usepackage{amsmath,amsfonts}
\usepackage{adjustbox}
\usepackage{algpseudocode}
\usepackage{array}
\usepackage[caption=false,font=normalsize,labelfont=sf,textfont=sf]{subfig}
\usepackage{textcomp}
\usepackage{stfloats}
\usepackage{url}
\usepackage{verbatim}
\usepackage{graphicx}
\usepackage{cite}
\usepackage[ruled,vlined]{algorithm2e}
\usepackage[table]{xcolor}  
\usepackage{multirow}  
\usepackage{longtable} 
\usepackage{booktabs}
\begin{document}

\title{Compensating Visual Insufficiency \\with Stratified Language Guidance \\for Long-Tail Class Incremental Learning}

\author{Xi Wang, Xu Yang, \textit{Member, IEEE}, Donghao Sun, Cheng Deng, \textit{Senior Member, IEEE}
\thanks{
Corresponding author: Cheng Deng.

Xi Wang, Xu Yang, Donghao Sun, and Cheng Deng are with the School of Electronic Engineering, Xidian University, Xi' an 710071, China (e-mail: \{wangxi6317, xuyang.xd, donghaosun508, chdeng.xd\}@gmail.com).}}

\markboth{IEEE TRANSACTIONS ON PATTERN ANALYSIS AND MACHINE INTELLIGENCE}%
{Shell \MakeLowercase{\textit{et al.}}: A Sample Article Using IEEEtran.cls for IEEE Journals}


\maketitle

\begin{abstract}
Long-tail class incremental learning (LT-CIL) remains highly challenging because the scarcity of samples in tail classes not only hampers their learning but also exacerbates catastrophic forgetting under continuously evolving and imbalanced data distributions. To tackle these issues, we exploit the informativeness and scalability of language knowledge. Specifically, we analyze the LT-CIL data distribution to guide large language models (LLMs) in generating a stratified language tree that hierarchically organizes semantic information from coarse- to fine-grained granularity. Building upon this structure, we introduce stratified adaptive language guidance, which leverages learnable weights to merge multi-scale semantic representations, thereby enabling dynamic supervisory adjustment for tail classes and alleviating the impact of data imbalance. Furthermore, we introduce stratified alignment language guidance, which exploits the structural stability of the language tree to constrain optimization and reinforce semantic–visual alignment, thereby alleviating catastrophic forgetting. Extensive experiments on multiple benchmarks demonstrate that our method achieves state-of-the-art performance.
\end{abstract}


\section{Introduction}

\IEEEPARstart{R}{eal} world data is inherently dynamic and imbalanced, leading to increased interest in long-tail class incremental learning (LT-CIL) \cite{wang2024long}. In LT-CIL, training tends to be dominated by head classes with many samples, leading to poor performance for tail classes with insufficient visual information. This disparity intensifies along dynamically evolving data streams, ultimately exerting a substantial negative impact on overall model performance.

\begin{figure}[t]
\centering
\includegraphics[width=1\columnwidth]{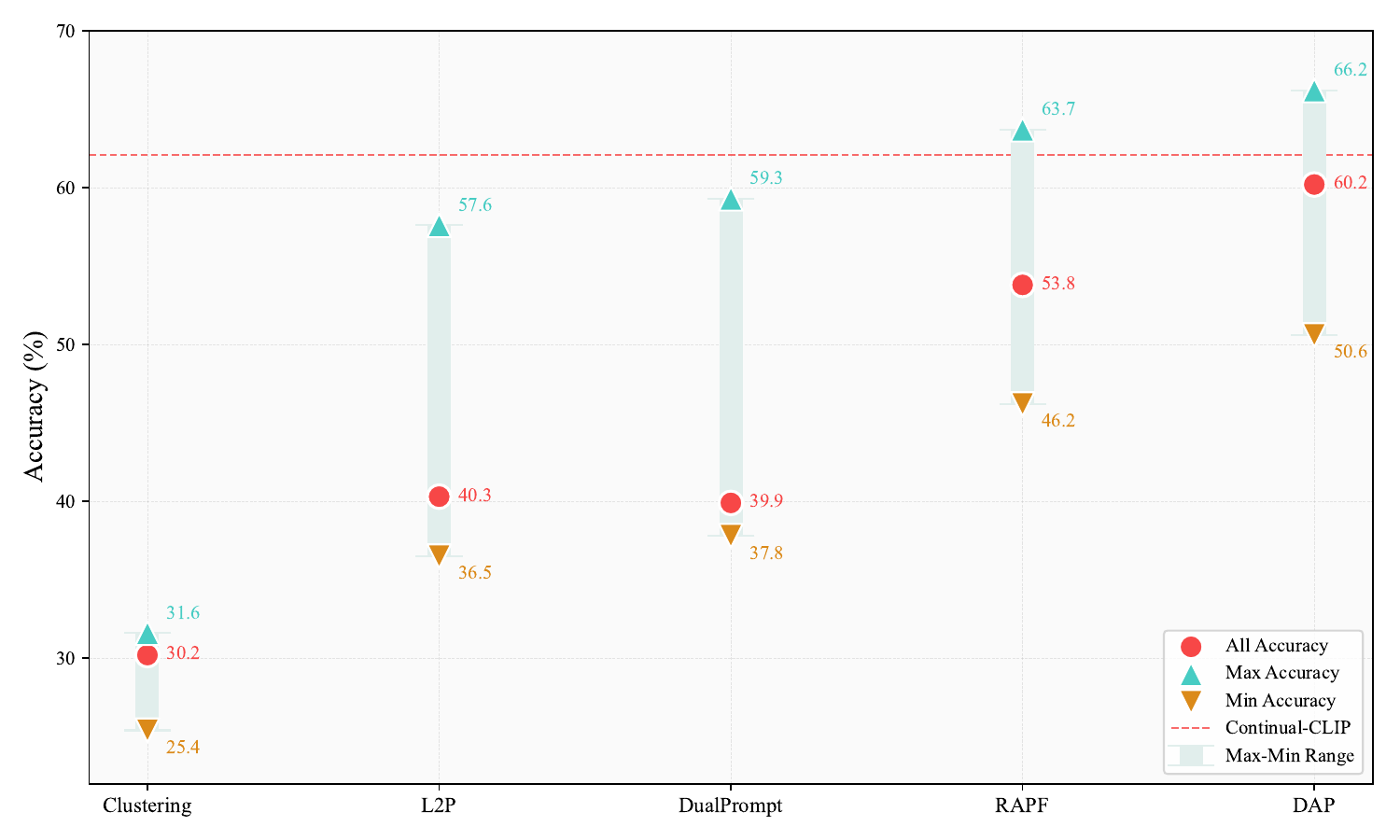} 
\caption{Experiments were conducted on CIFAR100, imbalance rate $\rho=0.01$ and 10 tasks. We evaluated accuracy across head classes, tail classes, and the complete dataset to assess the influence of semantic information on LT-CIL. }
\label{fig1}
\end{figure}

Most existing LT-CIL methods are adapted from class incremental learning (CIL) methods \cite{DBLP:conf/aaai/ZhaiL0C24}, including data re-balancing strategies \cite{liu2022long, kalla2024robust}, or parameter-efficient fine-tuning (PEFT) of pre-trained models (PTMs) \cite{gu2025dynamic, qi2025adaptive}. Recently, language-guided methods has received much attention in CIL \cite{DBLP:journals/corr/abs-2210-03114, khan2023introducing, park2024pre, zhao2024ltgc}, which typically utilizes fixed templates, such as \textit{'a photo of [class]'}, to compute semantic–visual relevance between text and image embeddings. We therefore explore the potential of language knowledge in addressing LT-CIL.

We conducted a preliminary validation on CIFAR100 under severe imbalance. Using the same backbone, we compared L2P \cite{wang2022learning}, DualPrompt \cite{wang2022dualprompt}, and DAP \cite{gu2025dynamic} (visual-only), as well as RAPF \cite{huang2025class}, (visual + semantic). L2P, DualPrompt, and RAPF are designed for CIL, whereas DAP specifically targets LT-CIL, we also included a zero-shot experiment using CLIP (Continual-CLIP) \cite{DBLP:journals/corr/abs-2210-03114} and a clustering experiment employing only the visual encoder (Clustering). We measured accuracy for head classes, tail classes, and all classes.  The experimental results in Fig.\ref{fig1} show that 1) overall performance after fine-tuning may be even lower than that of the PTMs itself, owing to the degradation performance deficiency in tail classes; and 2) among existing CIL methods, those incorporating semantic information are less susceptible to imbalanced, continuous data; These findings suggest that semantic information provides a promising direction for mitigating the challenges of LT-CIL. However, fixed templates are limited and hinder the full exploitation of semantic cues. Therefore, this paper aims to explore how semantic information can be effectively leveraged to overcome the key limitations of LT-CIL.

In this work, we first analyze the distributional characteristics of LT-CIL and design a recursive algorithm with four customized prompt templates to guide large language models (LLMs) in generating task-specific text descriptions. Following a hierarchical order from coarse to fine granularity, these texts are organized into a task-specific stratified language tree (SL-Tree). Based on the tree, stratified adaptive language guidance leverages learnable weights to integrate multi-scale semantic information, thereby enabling dynamic adjustment of the supervisory signal for tail classes with limited visual information and mitigating the impact of data imbalance. In addition, by exploiting the structural stability of the SL-Tree during training, stratified alignment language guidance constrains model optimization through the relative similarity distributions between the semantic and visual modalities, thereby further alleviating catastrophic forgetting. With two parallel stratified language guidances, we significantly enhance the PTM performance on LT-CIL. Our main contributions can be summarized as follows:
\begin{itemize}
    \item We introduce a stratified language tree, generated by LLMs, to organize multi-scale semantic information and provide structured supervision.
    \item We propose leveraging the rich semantic information and stability of the stratified language tree to guide model training in the visual space, thereby alleviating catastrophic forgetting exacerbated by imbalanced data.
    \item We perform extensive experiments to demonstrate the effectiveness of our method, all achieving state-of-the-art results.
\end{itemize}

\section{Related Works}

\begin{figure*}[t]
\centering
\includegraphics[width=1.8\columnwidth]{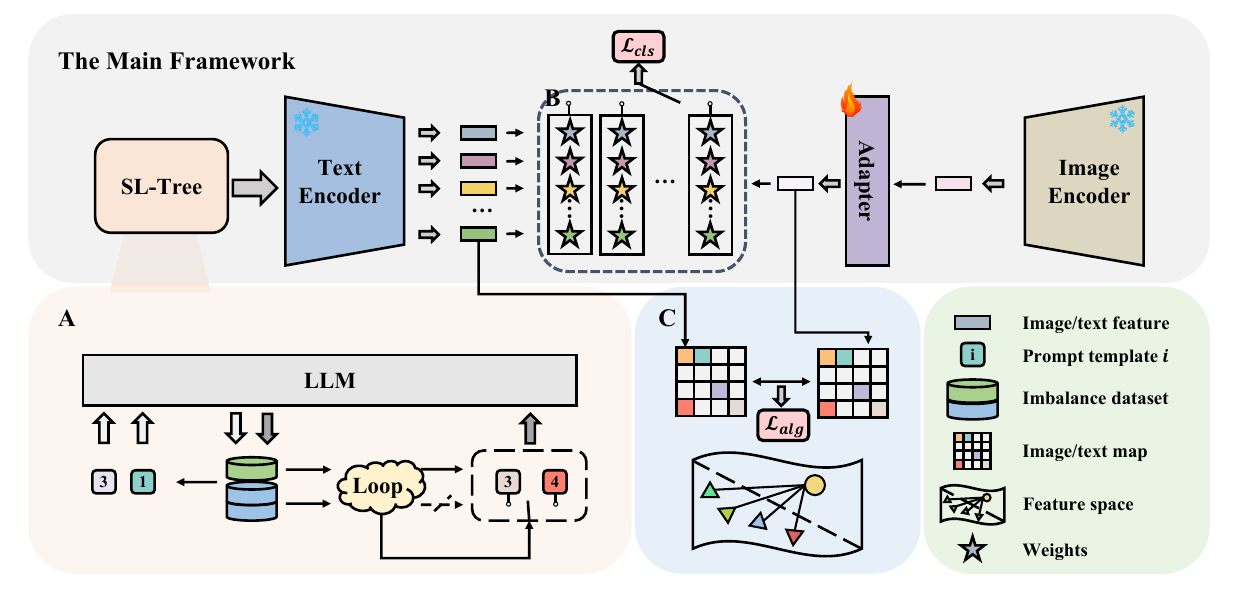} 
\caption{The overall framework of our method. A is the stratified language tree generation, B is the stratified adaptive language guidance, and C is the stratified alignment language guidance.}
\label{fig2}
\end{figure*}

\subsection{Class Incremental Learning with Pre-trained Models}
Class incremental learning (CIL) aims to continually acquire new knowledge from a non-stationary data stream continuously \cite{aljundi2018memory,lee2017overcoming,ritter2018online,zenke2017continual,DBLP:journals/pami/MazumderSRN24,DBLP:journals/pami/LiWQLL25,DBLP:journals/pami/ZhuZCL25,DBLP:journals/pami/ZhaoLST25, DBLP:conf/cvpr/FiniCA0AM22, DBLP:conf/cvpr/LinCL22, DBLP:conf/iclr/LangeVT23, DBLP:conf/iclr/0001NGC23, DBLP:conf/iclr/WangL0H24}. The primary challenge is learning without catastrophic forgetting: as new data arrives, the model's performance on previously learned tasks should not significantly degrade \cite{li2017learning, CIL0104, DBLP:conf/cvpr/Gu0WD22, DBLP:conf/ijcai/WeiDY20, DBLP:conf/iclr/QiaoZTCQP024}. Recent advances in CIL that leverage pre-trained models have provided promising avenues for balancing generalization and adaptability \cite{DBLP:conf/wacv/LeeZW23}, thus mitigating catastrophic forgetting. Prompt-based methods \cite{wang2022learning, wang2022dualprompt, smith2023coda} have demonstrated the effectiveness of adapting pre-trained models into continuous data streams. A two-stage method \cite{wu2022class}, utilizing feature augmentation and classifier fusion, has shown improvements for CIL. Some methods \cite{khan2023introducing, cao2024generative,huang2025class} incorporate knowledge from the language modality to assist model learning, and these methods have been successful. Collectively, these strategies underscore the potential of integrating pre-trained models within CIL, enhancing model performance. 

\subsection{Long-tail Class Incremental Learning}
Long-tail class incremental learning (LT-CIL) addresses the challenge of data imbalance \cite{zhang2023deep, li2022trustworthy} and catastrophic forgetting in a continuously evolving data stream. Methods such as LUCIR \cite{liu2022long} and GVAlign \cite{kalla2024robust} focus on balancing class distributions through regularization and distance constraints to enhance tail class learning. ISPC \cite{wang2024long} introduces the independent sub-prototype space and reminiscence space to tackle data imbalance and catastrophic forgetting simultaneously. Recently, several methods have been proposed based on parameter-efficient fine-tuning (PEFT) using pre-trained models \cite{gu2025dynamic, qi2025adaptive}. 

However, most existing methods still struggle to fully exploit the semantic structure of the data, particularly for underrepresented tail classes, leaving room for improvement in knowledge transfer and generalization.

\section{Method}
Our objective is to enable the network to sequentially learn multiple tasks from imbalanced data streams. In this section, we present the problem definition of LT-CIL, followed by a detailed description of the proposed method.

\subsection{Preliminary}
\subsubsection{Problem definition}
Typically, we consider a supervised LT-CIL setting, where a model needs to consecutively learn $T$ different tasks. Each task $t$ contains different classes ${C}^t$  and there is no overlap between any two different tasks: ${C}^i \cap {C}^j =\emptyset$, for $i \ne j$, and $(x, y) \in \mathcal{D}^{t}$ denotes a training sample in task $t$. We characterize the degree of data imbalance by the imbalance rate $\rho$, defined as the ratio between the largest and smallest sample sizes. After processing the dataset according to $\rho$, we partition it into different sub-tasks, with head and tail classes randomly assigned to each task, thus creating an imbalanced and continuous data flow.
\subsubsection{Pre-trained model}
We adopt CLIP \cite{radford2021learning} as the pre-trained model for our method due to its capability to jointly process visual and textual modalities. An efficient method for adapting a pre-trained model to downstream tasks involves integrating a lightweight network as an adapter. We denote the visual encoder as $E_{v}$, the text encoder as $E_{t}$, and the linear adapter as $f$. Given input $(x,y)$, $y$ denotes the text form of the label in all subsequent descriptions (e.g., cat). The working process of the model can be represented as:
\begin{align}
pred = arg\max_{i}\left(f\left(E_{v}(x)\right) \cdot  E_{t}(y_i)\right).
\label{eq1}
\end{align}

We argue that the fine-tuning strategy of the adapter is inherently influenced by continuous, imbalanced data. As discussed before, tail classes suffer from insufficient visual samples, which makes it difficult for the model to learn robust representations. To address this, we expect semantic information to provide a stronger supervisory signal for tail classes. Accordingly, we first generate task-specific text descriptions.
\subsection{Stratified Language Tree}
Specifically, we design four distinct prompt templates and a recursive algorithm to guide and constrain the LLM in generating the SL-Tree $G$ containing rich, structured text information.

For each task $t$ arriving in chronological order, we collect labels $y$ of all classes within the task and generate initial text description according to a \textbf{Fixed Template}, \textit{`a photo of [$y_i$]'}, which forms the base layer of the SL-Tree, $G^1$. 

Building upon this, to simulate the coarse categorization of objects, we design \textbf{Prompt Template 1}, \textit{`Please summarize the task in one sentence from the point of view of category which includes both [$y_1$ + $y_2$ + ...]'}, where [ $y_1$ + $y_2$ + ...] is the concatenation of all class labels within the task. With prompt 1, the LLM generates a task-level text description, primarily capturing common attributes shared by all classes within the task. This coarse-grained description constitutes the first level of the SL-Tree, $G^0$. 
\begin{figure}[h]
\centering
\includegraphics[width=0.95\columnwidth]{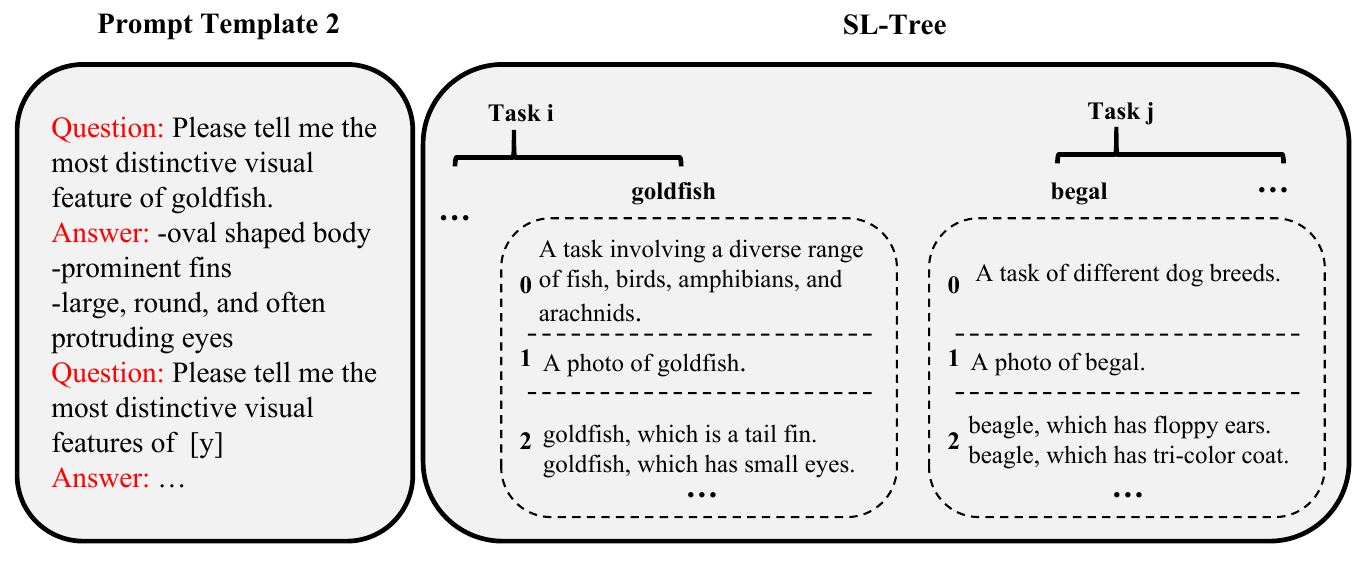} 
\caption{Illustration of Prompt Template 2 and SL-Tree.}
\label{fig3}
\end{figure}

Furthermore, fine-grained text descriptions that provide more granular details are required. To achieve this, we designed \textbf{Prompt Template 2}, \textit{`Please tell me the most distinctive visual feature of [$y_i$]'}, where [$y_i$] is sequentially replaced with the class labels. Guided by prompt 2, the LLM generates detailed descriptions for each class, highlighting distinctive features, such as color, pattern, or shape. These form the second level of the SL-Tree $2^{nd}$, $G^2$. However, this text generation process remains generic and does not reflect the imbalanced data distribution, resulting in uniform descriptions across all classes without sufficient differentiation for tail classes.
\begin{algorithm}[t]
\small
\label{al1}
    \SetAlgoLined
    \caption{SL-Tree Generation.}
    \KwIn{Dataset $(x,y)$, LLM, 4 prompt templates, fix template.}
    \KwOut{SL-Tree $G$}
    $G^0 \leftarrow$ LLM($y$ + Prompt 1)\;
    $G^1 \leftarrow$ LLM($y$ + Fixed Template)\;
    $G^2 \leftarrow$ LLM($y$ + Prompt 2)\;
    $\mathcal{A} \leftarrow$ Eq\ref{eq2}-Eq\ref{eq4} on $G^{0:2}$\;
    \For{each $\mathcal{A}_j$}{
        index = 0\;
        \While{$len|\mathcal{A}_j| > 2$ \textbf{and} index $< 8$}{
            $G^{index+3} \leftarrow$ LLM($y_j$ + $G^0$ + Prompt 3)\;
            index += 1\;
            $\mathcal{A}_j \leftarrow$ Eq\ref{eq2}-Eq\ref{eq4} on $G^{0:index}$\;
        }
        \If{$len|\mathcal{A}_j| == 2$}{
           $G^{index} \leftarrow$ LLM($y_j$ + Prompt 4)\;
        }
    }
    \Return $G$
\end{algorithm}

To address this, we use text encoder $E_t$ to process the current SL-Tree. It should be noted that if the generated text exceeds a certain length, CLIP cannot process it. Therefore, we constrain the form of the generated texts. The responses are expressed as multiple phrases, and several short phrases are retained within the same layer. The prompt format and the generated SL-Tree structure are illustrated in Fig.\ref{fig3}. The obtained text representations $g_i$ corresponding to each class can be represented as:
\begin{equation}
g_i = \frac{1}{L} \sum_{l=1}^{L} \frac{1}{N_i^{l}}\sum_{j=1}^{N_i^{l}} E_t(G_i^l[j]),
\label{eq2}
\end{equation}
where $l$ denotes the total number of layers of the SL-Tree and $N_i^{l}$ denotes the number of text descriptions in the corresponding layer. Next, we calculate the similarity between each tail classes and all other classes in the same task to construct the similarity matrix $Q_j$ for a tail classes $j$:
\begin{equation}
Q_j = cos(g_j,g_k) = [q_{jk}]_{j\in\left |  C^{min}\right |,k\in\left | C^t \right |},
\label{eq3}
\end{equation}
and identify all classes similar to tail class $j$, defining them as a confusion cluster $\mathcal{A}_{j}$,
\begin{equation}
\mathcal{A}_{j} = \left \{ \left ( j,k \right )|1-q_{jk}<  0.5 \right \}.
\label{eq4}
\end{equation}

If more than two classes are found to be similar to the central tail class $j$ in the cluster $\mathcal{A}_{j}$, i.e, $len(\mathcal{A}_{j}) \geq 3$, we introduce \textbf{Prompt Template 3}, \textit{`Please tell me the most distinctive visual features of [$y_i$] from the datasets which include [task description]'}. Here, [$y_i$] is iteratively replaced with the labels of all classes in confusion cluster, while [task description] is substituted with $G^{0}$. Using prompt 3, LLM generates text descriptions emphasizing distinctive inter-class differences at the task level.

Then we recalculated equation (2)-(4) for classes in the confusion cluster $\mathcal{A}_{j}$, until only one similar class remains, i.e, $len(\mathcal{A}_{j}) = 2$. At this stage, we introduce \textbf{Prompt Template 4}, \textit{`Please tell me the most distinctive visual features of [$y_i$] compared to [$y_j$]'}. Prompt 4 guides the LLM to perform one-to-one comparisons between tail class and the most similar class, generating highly discriminative comparative descriptions. The text generated with Prompt 4 constitutes the final layer, $G^{L}$, while those generated by Prompt 3 are inserted into the intermediate layers. However, real-world comparisons can be more complex than anticipated, and the number of classes in the confusion cluster may remain greater than 2 even after multiple iterations. Therefore, we set a maximum iteration limit of nine to prevent infinite looping. The overall generation process of the SL-Tree is detailed in Algorithm 1.

Throughout this recursive process, we obtain an SL-Tree specific to each sub-task, which follows the expected progression from coarse- to fine-grained text representation, and we merge the newly generated SL-Tree and existing one. Then, we utilized its rich knowledge and stability to guide visual space learning and mitigate the challenges of LT-CIL.

\subsection{Stratified Adaptive Language Guidance}
In the SL-Tree, tail classes are assigned more text descriptions to compensate for their limited visual samples. During training, we aim to fully exploit these texts to provide more refined supervisory signals for tail classes while avoiding interference with the learning of other classes. Motivated by this, we propose stratified adaptive language guidance.

Specifically, we process the SL-Tree using text encoder $E_t$, and the corresponding text features of each layer $l$ can be represented as follows:
\begin{equation}
\begin{aligned}
&g^l = \left[g^l_1,g^l_2,\dots ,g^l_i, \dots\right]_{i\in\left | C^{1:t} \right |,l \in L},\\
&g^l_i = 
\left\{
\begin{matrix}

&\frac{1}{N_i^{l}}\sum_{j=1}^{N_i^{l}}E_t(G_i^l[j]) & \text{if} \quad \text{len}(G_i^l) > 0. \\
& 0 & \text{else}  

\end{matrix}
\right.
\end{aligned}
\label{eq5}
\end{equation}

Equation (\ref{eq5}) indicates that for the $i^{th}$ class in the $l^{th}$ layer, if text descriptions are available, the mean of their text features are calculated. otherwise, the value is set to $0$, implying that the class is not included in that round of comparison.

For any input $x$, the visual features are sequentially passed through the SL-Tree. The final prediction, aggregated across all layers, is formulated as
\begin{equation}
p(x) = \sum_{l = 0}^{L} a_l\left(f(E_{v}(x))\cdot g^l\right),
\end{equation}
where $\alpha_l$ denotes the importance weight of each layer.

Different layers in the SL-Tree contribute unequally to the final prediction; thus, the weights $\alpha_l$ should be adaptively adjusted. We treat them as learnable parameters and embed them into the optimization objective ${ \mathcal { L } } ( \theta , \alpha )$, where $ \mathcal{ L }$ is the cross entropy loss, i.e., $\mathcal{L} = \mathcal{L}_{ce}(p(x), y) $. 

During training, updates of $\alpha$ and $\theta$ are performed alternately. The network parameters $\theta$ are optimized via standard stochastic gradient descent. However, $\alpha$ acts as a set of weights to integrate multi-scale semantic information, which requires a more delicate optimization objective.

To prevent weight collapse into a single layer, we introduce the negative entropy of the weights as a  regularization constraint,
\begin{equation}
R _ { { con } } ( \alpha ) = \sum _ { l = 0 } ^ { L } \alpha _ { l} \, \log ( \alpha _ { l } + \varepsilon ),
\end{equation}
where $\varepsilon = 1 \times 10^{-8}$ prevents numerical instability. $R_{{con}}$ serves as a smoothness prior on the layer-wise importance weights. By maximizing the entropy of $\alpha$, the model avoids degenerating into a single-layer reliance, encouraging distributed semantic aggregation.

Moreover, the update of $\alpha$ is also affected by the imbalanced distribution. Accordingly, each class is associated with its own set of weights, $\alpha = \left\{\alpha_{0,c}, \alpha_{1,c}, \cdots \alpha_{L,c}\right\}_{c=0}^{C^{1:T}} $, and all sets of weights updated simultaneously. 

For tail classes that are more difficult to distinguish, theoretically more comparisons and queries will be performed, and more fine-grained text descriptions will be assigned in SL-Tree. When integrating predictions across layers, we expect tail classes to rely more on high-level semantic decisions. Therefore, we propose a frequency prior constraint for tail classes to enhance their discriminability. We first define the prior distribution,
\begin{equation}
\pi_{l,c} = \exp\bigl( \kappa_c \cdot \phi_l \bigr), \quad \kappa_c = \left( \frac{ \bar{n} }{ {n_{{c}}} } \right),
\end{equation}
where $\bar { n }$, ${n_{{c}}}$ denote the average number of all samples and the number of samples for  class $c$ in the current task and $\phi _ { l } = ( l - 1 ) / ( L - 1 )$. To enforce high-level priority, we apply KL divergence as a regularization term:
\begin{equation}
{R} _ { { f r e q } }(\alpha) = \sum_{k } { K L } \big ( \alpha_{:,k}\; \big | \big | \; \pi _ { :,  k } \big ) .
\end{equation}
${R} _ { { f r e q } }$ introduces a data-dependent prior that rebalances the influence of coarse- and fine-grained layers according to class frequency. Intuitively, it forces tail classes to rely more on higher-level semantic abstractions, mitigating the insufficient visual cues.

In summary, the complete optimization objective for both parameters during training are
\begin{equation}
\left\{
\begin{aligned}
&\min_{\alpha}\Big[\mathcal{L}_{ce}(\theta,\alpha) + R_{con}(\alpha) + R_{freq}(\alpha)\Big] \\[2pt]
&\min_{\theta}\mathcal{L}_{ce}(\theta,\alpha)
\end{aligned}
\right.
\label{eq:alpha_theta_optimization}
\end{equation}

It should be noted that $\alpha$ represents the weights of different layers, which must be satisfied $\sum_{l}\alpha_{l,c}=1$ and their values lie on the probability simplex \cite{DBLP:journals/corr/WangC13a}. Therefore, the update of $\alpha$ must be constrained to the $(L-1)$-dimensional simplex:
\begin{equation}
\begin{aligned}
\alpha ^ { ( m + 1 ) }_{l,c} = \Pi _ { \Delta } \Big ( \alpha ^ { ( m ) }_{l,c} - \eta _ { \alpha } \, \nabla _ { \alpha } \mathcal { L } ( \theta , \alpha ^ { ( m ) }_{l,c} ) \Big ) \, , \\
\Pi _ { \Delta } = \Big \{ \alpha _ { l,c } \! \geq \! 0 , \sum _ { l } \alpha _ { l,c } \! = \! 1 \Big \} .
\end{aligned}
\end{equation}
Where $m$ denotes training steps, and $\eta _ { \alpha }$ is learning rate. $\Pi _ { \Delta }$ denotes to the closed-form simplex projection \cite{DBLP:conf/icml/DuchiSSC08}. Specifically, $v_i = \alpha ^ { ( m ) }_{i,c} - \eta _ { \alpha } \, \nabla _ { \alpha } \mathcal { L } ( \theta , \alpha ^ { ( m ) } _{i,c})$. After sorting the elements of vector $v$ in descending order, $v{(0)} \ge v_{(1)} \ge \dots \ge v_{(L)}$, we determine the cutoff point:
\begin{equation}
i ^ { \star } = \operatorname* { m a x } \left\{ i \in \{ 0 , \dots , L \} : v _ { ( i ) } - \frac { 1 } { i } \Big ( \sum _ { j = 0 } ^ { i } v _ { ( j ) } - 1 \Big ) > 0 \right\}.
\end{equation}
Then, the threshold $\tau$ can be calculated as
\begin{equation}
\tau = \frac { 1 } { i ^ { \star } } \Big ( \sum _ { i = 0 } ^ { i ^ { \star } } v _ { ( i ) } - 1 \Big ),
\end{equation}
and the final closed-form solution is
\begin{equation}
\Pi _ { \Delta } ( v _i) = \operatorname* { m a x } ( v _ { i } - \tau , 0 ).
\end{equation}

Through the proposed stratified adaptive language guidance, tail classes receive stronger semantic supervision. However, during training, the model is still affected by catastrophic forgetting. Since the SL-Tree and the text encoder $E_t$ remain frozen and do not undergo parameter updates. Therefore, we assume that, in the absence of catastrophic forgetting, the relative distance between the semantic space and the visual space should remain stable. Building on this observation, we further propose stratified alignment language guidance.
\subsection{Stratified Alignment Language Guidance}
After each task, we store the prototype $\mu_c$, $\mu_c= \frac{1}{n_c} {\textstyle \sum_{i=1}^{n_c}}E_{v}(x_i)$ and covariance matrix $\Sigma_c$ of each class $c$, 
\begin{equation}
\Sigma _c = \frac{1}{n_c-1} (X_c-\bar{{X_c}})(X_c-\bar{{X_c}})^{T},
\end{equation}
where $n_c$ denotes the number of class $c$ within the batch, $X_c$ is the the matrix consisting of all features and $\bar{{X_c}}$ represents the mean of each feature dimension of classes $c$.

Specifically, for any task $t,t>0$, we first compute the frequency of occurrence for all classes within the mini-batch $\mathcal{B}$, and identify the highest frequency $r$. The value $r$ is then used as the sampling number to sample the features of each old class $k$ from the normal distribution $\mathcal{N}(\mu_k, \Sigma_k)$. The sampled feature batch $\mathcal{B}_{sam}$ is subsequently mixed with the original one to form a new approximate balanced batch $\mathcal{B}_{bal} = cat(\mathcal{B},\mathcal{B}_{sam})$,  where $cat$ denotes concatenation. We then compute pairwise similarities between the sample points within the batch to obtain a visual batch similarity matrix,
\begin{equation}
\mathcal{S}_{v} = \Psi (f(\mathcal{B}_{bal})) \cdot (\Psi (f(\mathcal{B}_{bal})))^{T},
\label{eq10}
\end{equation}
where $\Psi $ denotes normalization. The resulting matrix $\mathcal{S}_{v}$ is treated as the visual-space distribution. Then, we compute the text features for class $c$ in $\mathcal{B}_{bal}$, $g_c = \sum_{l = 0}^{L} \alpha_{l,c} \cdot g_c^l$.

Similar to the visual space above, we can compute the similarity in the semantic space to obtain the semantic batch similarity matrix $\mathcal{S}_t$. Because the SL-Tree and text encoder remain frozen during training, we infer that the semantic space exhibits strong stability. Thus, we can use the distribution within the semantic space to constrain the optimize. Therefore, we compute the similarity between the distributions of the visual and the semantic space as follows,
\begin{equation}
       \mathcal{L}_{alg} = \frac{1}{2|\mathcal{B}_{bal}|} \sum_{|\mathcal{B}_{bal}|}^{}\left(\mathcal{S}_v\log\left ( \frac{\mathcal{S}_v}{\mathcal{S}_t}  \right ) 
+ \mathcal{S}_t\log\left ( \frac{\mathcal{S}_t}{\mathcal{S}_v}  \right)\right).
\label{eq12} 
\end{equation}

To further maximize the performance of our method, we compute the distillation loss between the old and new models, $\mathcal{L}_{kd} = \left \| f_{old}(E_{v}(x))-f_{new}(E_{v}(x))\right \| _{2}$. Thus, the overall training objective in equation (10) becomes
\begin{equation}
\left\{
\begin{aligned}
\min_{\alpha}\ &
\Big[
\mathcal{L}_{ce}(\theta,\alpha)
+ \lambda_1 \mathcal{L}_{alg}(\theta,\alpha)
+ \lambda_2 \mathcal{L}_{kd}(\theta,\alpha) \\[3pt]
&\quad
+ \lambda_3 R_{{con}}(\alpha)
+ \lambda_4 R_{{freq}}(\alpha_{:,t})
\Big] \\[4pt]
\min_{\theta}\ &
\Big[
\mathcal{L}_{ce}(\theta,\alpha)
+ \lambda_1 \mathcal{L}_{alg}(\theta,\alpha)
+ \lambda_2 \mathcal{L}_{kd}(\theta,\alpha)
\Big]
\end{aligned}
\right.
\label{eq:merged_loss}
\end{equation}

In the inference phase, for any input $x$, we calculate the prediction results $p_{\alpha_{:,c}}(x)$ under different parameters. We then define the difference between the maximum and the second maximum of the predicted results as the decision margin $m_{\alpha}(x)$,
\begin{equation}
m_{\alpha_{:,c}}(x) = arg\max_i p_{\alpha_{:,c}}(x)[i] - arg\max_{i' \neq i} p_{\alpha_{:,c}}(x)[i'].
\end{equation}
Finally, we select the prediction with the larger decision margin as the final prediction.

\subsection{Parameters Independence Analysis}
Equation (18) integrates complementary constraints: $\mathcal{L}_{ce}$ enforces discriminability, $\mathcal{L}_{kd}$ ensures knowledge retention, $\mathcal{L}_{alg}$ maintains inter-modal consistency, and the two regularizers $R_{con}$ and $R_{freq}$ stabilize the adaptive weight distribution. In our experiments, the updates of $\theta$ and $\alpha$ are performed alternately. During the derivation of $\theta$, $\alpha$ is treated as a constant, and vice versa. Therefore, the optimization of the two parameter sets does not interfere with each other.

Moreover, since $\alpha$ consists of differnet sets of parameters, we also explain the independence between different components of $\alpha$. As a first step, we reorganize equation (18) with respect to each class.
\begin{equation}
\label{eq:combined_loss}
\begin{aligned}
\mathcal{L}(\theta,\alpha) 
&= \sum_{k = 0}^{C^{1:T}} \mathcal{J}_{k}(\theta,\alpha_{:,k}), \\[3pt]
\mathcal{J}_{k}(\theta,\alpha_{:,k}) 
&= \sum_{(x,y)\in\mathcal{D}_k}
   \Big( \mathcal{L}_{{ce}}
       + \lambda_1 \mathcal{L}_{{alg}}
       + \lambda_2 \mathcal{L}_{{kd}} \Big)(\theta,\alpha_{:,k}) \\
&\quad + \lambda_3\,R_{{ent}}(\alpha_{:,k}) + \lambda_4\,R_{{freq}}(\alpha_{:,k}).
\end{aligned}
\end{equation}
We then obtain the derivatives for different subsets of $\alpha$,
\begin{equation}
\begin{cases}
\nabla_{\alpha_{:,k}} \mathcal{L} = \nabla_{\alpha_{:,k}} \mathcal{J}_{k}, \\[3pt]
\dfrac{\partial^2 \mathcal{L}}{\partial \alpha_{:,k}\,\partial \alpha_{:,k'}} = 0, \quad k \ne k'.
\end{cases}
\end{equation}
This result indicates that the first-order derivative of the optimization objective with respect to each $\alpha$ depends only on its own parameters, while all cross-partial derivatives vanish. Thus, the different sets of adaptive weights are completely independent and can be optimized separately without mutual influence.

\section{Experiments}
\begin{table*}[t]
\centering
\caption{Comparison experiments on ImageNet-R, \textbf{bolded} indicates optimal, \underline{underlined} indicates sub-optimal.}
\begin{adjustbox}{width=0.8\textwidth}
\begin{tabular}{@{}lcccccccc@{}}
\toprule
\multirow{3}{*}{Method} & \multicolumn{4}{c}{$\rho = 0.1$} & \multicolumn{4}{c}{$\rho = 0.01$} \\ \cmidrule(l){2-9} 
 & \multicolumn{2}{c}{10 tasks} & \multicolumn{2}{c}{20 tasks} & \multicolumn{2}{c}{10 tasks} & \multicolumn{2}{c}{20 tasks} \\
 & $A_{last}\,(\uparrow) $ & $F_{avg}\,(\downarrow)$ \ &$A_{last}\,(\uparrow) $ & $F_{avg}\,(\downarrow)$ & $A_{last}\,(\uparrow) $ &$F_{avg}\,(\downarrow)$ & $A_{last}\,(\uparrow) $ & \multicolumn{1}{c}{$F_{avg}\,(\downarrow)$} \\ \midrule
LFM+MMS \cite{franklin2024text} & 24.7 & 26.7 & 22.7 & 29.1 & 16.8 & 27.1 & 15.7 & \multicolumn{1}{c}{28.9} \\ \midrule
PODNET + LWS \cite{liu2022long} & 49.7 & 6.3 & 49.5 & 6.0 & 48.3 & 6.5 & 47.6 & \multicolumn{1}{c}{6.1} \\
PODNET + GVAlign \cite{kalla2024robust} & 53.6 & 6.1 & 52.9 & 5.8 & 51.3 & 6.2 & 50.8 & \multicolumn{1}{c}{5.9} \\
ISPC \cite{wang2024long} & 57.0 & 5.6 & 55.9 & 5.9 & 52.1 & 5.5 & 51.8 & \multicolumn{1}{c}{5.3} \\ \midrule
PriViLege \cite{park2024pre} & 62.7 & 3.6 & 57.8 & 3.1 & 50.4 & 3.3 & 45.9 & \multicolumn{1}{c}{3.0} \\
L2P \cite{wang2022learning} & 67.6 & 4.8 & 60.1 & 3.3 & 50.9 & 5.4 & 47.6 & \multicolumn{1}{c}{5.9} \\
DualPrompt \cite{wang2022dualprompt} & 68.9 & 3.1 & 61.2 & 3.1 & 51.1 & 5.0 & 47.2 & \multicolumn{1}{c}{5.6} \\
CODAPrompt \cite{smith2023coda} & 71.1 & 3.0 & 66.5 & 2.6 & 56.8 & 4.5 & 52.2 & \multicolumn{1}{c}{4.6} \\
GMM \cite{cao2024generative} & 71.2 & 2.1 & 65.7 & 2.1 & 62.3 & 2.3 & 60.7 & \multicolumn{1}{c}{\underline{2.2}} \\
RAPF \cite{huang2025class} & 72.0 & 1.9 & 65.8 & 2.0 & 63.5 & 2.1 & 61.3 & \multicolumn{1}{c}{2.4} \\ 
MG-CLIP \cite{DBLP:journals/corr/abs-2507-09118} & \underline{72.2} & 1.7 & 68.2 & \underline{1.6} & 64.9 & 2.0 & 62.9 & \multicolumn{1}{c}{2.5} \\ 
\midrule
DAP \cite{gu2025dynamic} & 71.4 & 1.9 & 70.0 & 2.0 & 64.8 & 2.1 & 61.9 & \multicolumn{1}{c}{2.4} \\ 
APART \cite{qi2025adaptive} & 71.8 & \underline{1.5} & \underline{70.3} & 1.7 & \underline{65.2} & \underline{1.9} & \underline{63.0} & \multicolumn{1}{c}{2.5} \\ 
\midrule
\rowcolor{gray!20}
\textbf{Ours }& \textbf{76.1} & \textbf{1.6} & \textbf{73.7} & \textbf{1.2} & \textbf{72.1} & \textbf{1.9} & \textbf{70.0}   & \multicolumn{1}{c}{\textbf{1.8}} \\ \bottomrule
\end{tabular}
\end{adjustbox}
\label{tab1}
\end{table*}

\begin{table*}[t]
\centering
\caption{Comparison experiments on CIFAR100, \textbf{bolded} indicates optimal, \underline{underlined} indicates sub-optimal.}
\begin{adjustbox}{width=0.8\textwidth}
\begin{tabular}{@{}lcccccccc@{}}
\toprule
\multirow{3}{*}{Method} & \multicolumn{4}{c}{$\rho = 0.1$} & \multicolumn{4}{c}{$\rho = 0.01$} \\ \cmidrule(l){2-9} 
 & \multicolumn{2}{c}{5 tasks} & \multicolumn{2}{c}{10 tasks} & \multicolumn{2}{c}{5 tasks} & \multicolumn{2}{c}{10 tasks} \\
 & $A_{last}\,(\uparrow) $ & $F_{avg}\,(\downarrow)$& $A_{last}\,(\uparrow) $ & $F_{avg}\,(\downarrow)$ & $A_{last}\,(\uparrow) $ & $F_{avg}\,(\downarrow)$ & $A_{last}\,(\uparrow) $ & $F_{avg}\,(\downarrow)$ \\ \midrule
LFM+MMS \cite{franklin2024text} & 39.7 & 22.6 & 27.4 & 25.7 & 30.6 & 25.1 & 19.8 & 21.0 \\ \midrule
PODNET + LWS \cite{liu2022long} & 51.9 & 5.1 & 51.0 & 4.3 & 35.9 & 6.1 & 35.2 & 6.0 \\
PODNET + GVAlign \cite{kalla2024robust} & 53.0 & 4.7 & 51.9 & 4.0 & 39.2 & 5.7 & 38.6 & 5.8 \\
ISPC \cite{wang2024long} & 53.4 & 4.1 & 52.4 & 3.8 & 39.7 & 5.9 & 39.0 & 5.6 \\ \midrule
PriViLege \cite{park2024pre} & 66.6 & 2.3 & 60.4 & 2.1 & 48.3 & 2.3 & 39.1 & 2.5 \\
L2P \cite{wang2022learning} & 67.5 & 2.0 & 61.3 & 2.5 & 48.9 & 2.2  & 40.3 & 2.7 \\
DualPrompt \cite{wang2022dualprompt} & 68.2 & 1.8 & 61.3 & 2.0 & 49.5 & 2.4 & 39.9 & 2.3 \\
CODAPrompt \cite{smith2023coda} & 74.4 & 1.6 & 69.2 & 1.9 & 61.7 & 1.8 & 51.9 & 2.4 \\
GMM \cite{cao2024generative} & 74.5 & 2.1 & 70.2 & 2.1 & 60.4 & 1.7 &  53.4&1.6 \\
RAPF \cite{huang2025class} & 75.4 & 1.9 & 70.9 & 1.9 & 62.5 & 2.0 & 53.8 & 1.6 \\
MG-CLIP \cite{DBLP:journals/corr/abs-2507-09118} & \underline{75.6} & 1.7 & \underline{71.3} & 1.8 & 62.9 & 1.9 & 54.5 & 1.6\\ \midrule
DAP \cite{gu2025dynamic} & 74.0 & 2.1 & 70.8 & 2.1 & 63.0 & 1.8 &  60.2&1.5 \\
APART \cite{qi2025adaptive} & 74.3 & \underline{1.5} & 71.0 & \underline{1.4} & \underline{63.3} & \underline{1.7} & \underline{60.9} & \textbf{1.4} \\
\midrule
\rowcolor{gray!20}
\textbf{Ours} & \textbf{77.3} & \textbf{1.3}& \textbf{72.0} & \textbf{1.2} & \textbf{64.3} & \textbf{1.8} & \textbf{64.0} & \underline{1.5}\\ \bottomrule
\end{tabular}
\end{adjustbox}
\label{tab2}
\end{table*}

\subsection{Experiments Setttings}
\textbf{Datasets.} We conduct our experiments using different benchmarks: CIFAR100 \cite{krizhevsky2009learning}, ImageNet-R \cite{DBLP:conf/iccv/HendrycksBMKWDD21} and CUB200 \cite{wah2011caltech}. We divided the ImageNet-R and CUB200 into $10$ or $20$ consecutive tasks of equal size, and CIFAR100 into $5$ or $10$ tasks. For CIFAR100, we applied imbalance rates $\rho = 0.1 $ and $\rho = 0.01 $ to simulate distribution imbalances. ImageNet-R is inherently imbalanced, with an approximate imbalance rate of $0.1$; we additionally construct a version with $\rho = 0.01$ for comparison. For CIFAR100 and ImageNet-R, we consider classes with fewer than 100 samples as tail classes, and the remaining ones as head classes. For CUB200, due to its smaller per-class sample size, we only consider $\rho = 0.1$, and treat classes with fewer than 10 samples as tail classes.

\textbf{Metrics.}
We use the standard metrics in continual learning to measure performance: {Last Accuracy: $A_{last}$, which calculates all seen classes’ accuracy after training for all tasks and Forgetting Rate: $F_{avg}$, which calculates the average forgetting of prior task knowledge.

\textbf{Comparison methods.} We compare our method with state-of-the-art algorithms for LT-CIL, including LWS\cite{liu2022long}, GVAlign \cite{kalla2024robust}, ISPC \cite{wang2024long}, DAP \cite{gu2025dynamic} and APART \cite{qi2025adaptive}. And we also select pre-trained model-based methods for CIL, L2P \cite{wang2022learning}, DualPrompt \cite{wang2022dualprompt}, CODAPrompt \cite{smith2023coda}, GMM \cite{cao2024generative}, RAPF \cite{huang2025class} and MG-CLIP \cite{DBLP:journals/corr/abs-2507-09118}. Additionally, we select few-shot CIL method PriViLege \cite{park2024pre} and long-tail learning method LFM \cite{franklin2024text}. It is worth noting that, RAPF, GMM, PriViLege, LFM, and MG-CLIP also incorporate semantic information based on a pre-trained model.

\textbf{Implementation details.}
For both datasets, our pre-trained model is ViT-B/16 of CLIP from OpenAI, and we train the model with the Adam optimizer for 30 epochs, while $\alpha$ updated once every 5 epochs. Both sets of parameters use a learning rate of $1\times 10^{-3}$. And the LLM we used is gpt-3.5-turbo. All experiments are obtained by re-running on Python 3.8, PyTorch 2.0.1, and a single NVIDIA A6000 GPU. All comparison methods that require the pre-trained model use ViT-B/16 of CLIP from OpenAI, and others use ViT-B/16 without pre-training. In our experiments, $\lambda_1 = 0.025$, $\lambda_2 = 1$, $\lambda_3 =0.3$ and $\lambda_4 =0.6$.
\begin{table}[t]
\centering
\caption{Comparison experiments on CUB200, \textbf{bolded} indicates optimal, \underline{underlined} indicates sub-optimal.}
\begin{adjustbox}{width=0.42\textwidth}
\begin{tabular}{@{}lcccc@{}}
\toprule
\multirow{2}{*}{Method} & \multicolumn{2}{c}{10 tasks} & \multicolumn{2}{c}{20 tasks} \\
 & $A_{last}\,(\uparrow) $ & $F_{avg}\,(\downarrow)$ & $A_{last}\,(\uparrow) $ & $F_{avg}\,(\downarrow)$ \\ \midrule
ISPC \cite{wang2024long} & 30.2 & 6.7 & 26.1 & 6.1 \\ \midrule
PriViLege \cite{park2024pre} & 30.7 & 5.9 & 20.1 & 2.7 \\
L2P++ \cite{wang2022learning} & 32.1 & 3.8 & 20.7 & 3.4 \\
DualPrompt \cite{wang2022dualprompt} & 32.0 & 3.1 & 20.3 & 3.0 \\
CODAPrompt \cite{smith2023coda} & 34.4 & 3.5 & 21.3 & 3.1 \\
RAPF \cite{huang2025class} & 40.9 & 2.8 & 38.1 & 2.6 \\
MG-CLIP \cite{DBLP:journals/corr/abs-2507-09118} & 41.2 & 2.7 & 40.5 & 2.6 \\ 
\midrule
DAP \cite{gu2025dynamic} & 42.4 & 3.5 & 40.9 & 3.1 \\
APART \cite{qi2025adaptive} & \underline{43.5} & \underline{2.6} & \underline{42.3} & \underline{2.4} \\ 
\midrule
\rowcolor{gray!20}
\textbf{Ours }& \textbf{51.5} & \textbf{2.5} & \textbf{51.0} & \textbf{2.1} \\ \bottomrule
\end{tabular}
\end{adjustbox}
\label{tab3}
\end{table}
\subsection{Experimental Results}
We conducted experiments on various datasets under different settings, and the results of ImageNet-R, CIFAR100 and CUB200 are shown in TABLE \ref{tab1}, \ref{tab2} and \ref{tab3} separately. When using ImageNet-R as the benchmark, with $\rho=0.1$ (i.e., no additional processing) and dividing the dataset into 10 tasks, our method achieves $76.1\%$ accuracy, representing a $3.9\%$ improvement over MG-CLIP. When the number of tasks is increased to $20$, our method continues to perform robustly, achieving an accuracy of $73.7\%$, which is a $3.4\%$ improvement over the previous SOTA method. Furthermore, when  $\rho = 0.01$, our method achieves $72.1 \%$ accuracy after learning $10$ consecutive tasks, and $70.0\%$ accuracy after 20 tasks. Our method maintains higher accuracy when the benchmark is changed to CIFAR100 and fine-grained CUB200 under different imbalanced ratios and tasks. Similarly, our method consistently remained optimal in comparisons of forgetting rates in all experiments, except on CIFAR100, $\rho=0.01$, $10$ tasks, which was $0.1\%$ higher than the APART.
\subsection{Ablation Study}
In this section, we examine the effectiveness of each module within our proposed method. The experiments were conducted on different dataset with 10 tasks, and the results are shown in Table \ref{tab4}. Take Imagenet-R with $\rho=0.01$ as an example, Baseline denotes zero-shot and it achieves a notable accuracy of $68.2 \%$. When SL-Tree is introduced, taking the mean of all layers, an improvement of $0.8\%$ is observed. Next, we explored the impact of training an adapter. Without additional constraints, severe catastrophic forgetting occurs, resulting in only $22.4 \%$ accuracy. 
\begin{table}[t]
\centering
\caption{Ablation study on different dataset with 10 tasks. + SL-Tree denotes taking the mean of all text features, while + updated $\alpha$ denotes using adaptive weights.}
\begin{tabular}{@{}lc*{4}{c}@{}}
\toprule
\multirow{2}{*}{Method} &
\multicolumn{2}{c}{CIFAR100} &
\multicolumn{2}{c}{ImageNet-R} &
CUB200 \\
\cmidrule(lr){2-3}\cmidrule(lr){4-5}\cmidrule(l){6-6}
 & $\rho=0.1$ & $\rho=0.01$ & $\rho=0.1$ & $\rho=0.01$ & $\rho=0.1$ \\
\midrule
Baseline & 62.1 & 62.1 & 68.2 & 68.2 & 45.6 \\
+ SL-Tree & 62.5 & 62.5 & 69.0 & 69.0 & 47.9 \\
+ $\mathcal{L}_{cls}$ & 19.8 & 17.4 & 23.8   & 22.4 & 20.1 \\
+ $\mathcal{L}_{kd}$ & 64.7 & 60.9 & 69.5  & 65.8 & 46.4\\
+ $\mathcal{L}_{alg}$ & 67.4 & 62.4 &  72.9  & 69.4 & 48.7 \\
+ updated $\alpha$ & 69.2 & 63.2 & 74.4 & 71.1 &  50.1\\
+ $\mathcal{R}_{ent}$ & 71.3 & 63.7 &75.6 & 71.5 & 50.9 \\
+ $\mathcal{R}_{freq}$ & 72.0 & 64.0 & 76.1 & 72.1 & 51.5 \\
\bottomrule
\end{tabular}
\label{tab4}
\end{table}
Introducing traditional knowledge distillation (KD) improves performance to $65.8 \%$, yet it still struggles to balance stability and plasticity during training. To address this, we propose stratified alignment language guidance that leverages the stability of the semantic space to regulate the optimization, resulting in $69.4\%$. Further enhancement is achieved by introducing learnable weights and the adjustment raises performance to $71.1 \%$, confirming that different layers contribute differently to predictions.When uniform distribution constraints and prior constraints on tail classes were further incorporated into the weight update, the accuracy improved to $71.5\%$ and $72.1\%$, respectively. Consistent performance across other experiments with different datasets further validates the effectiveness of our method.
\begin{figure}[h]
\centering
\includegraphics[width=1\columnwidth]{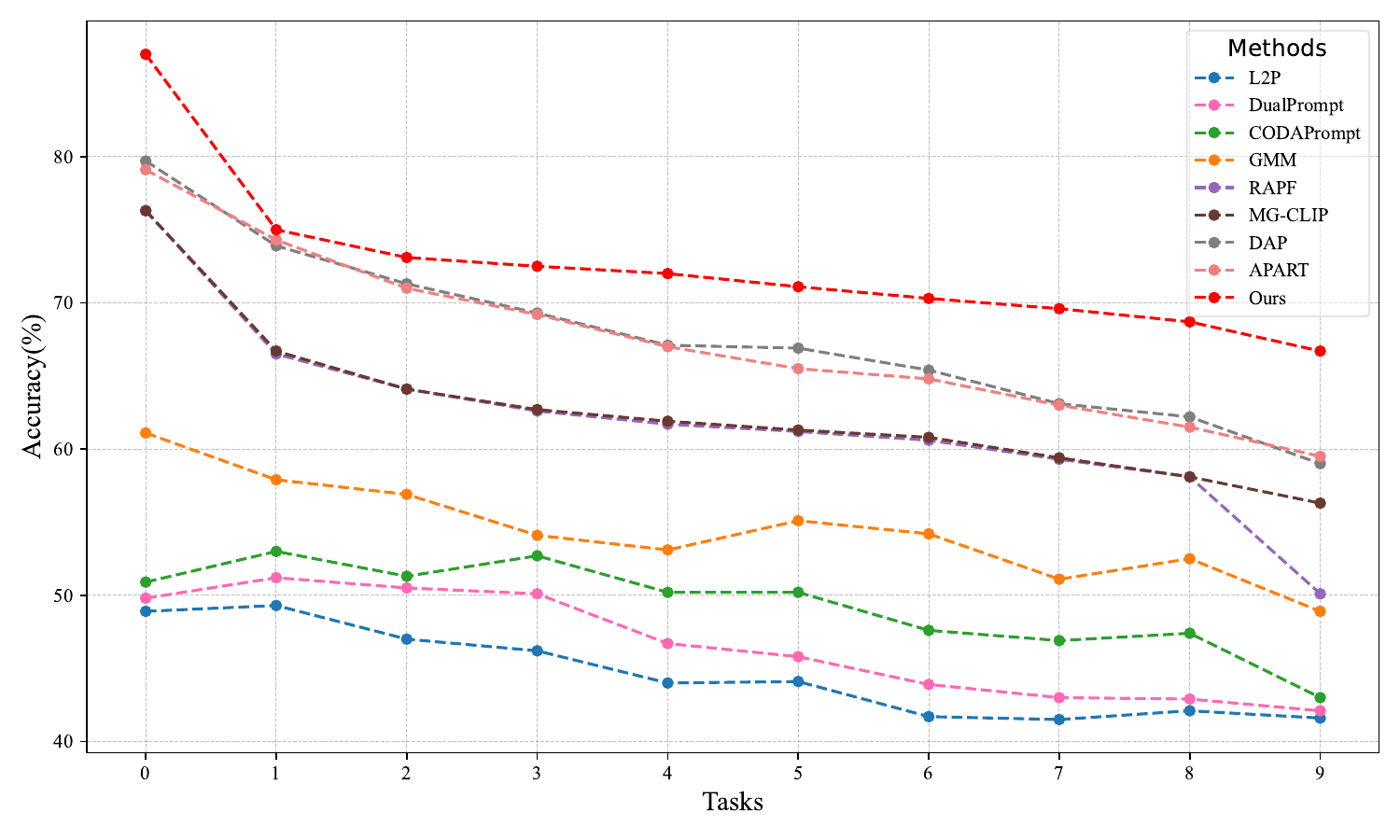} 
\caption{Tail classes accuracy of different tasks on ImageNet-R, $\rho=0.01$ after 10 tasks.}
\label{fig4}
\end{figure}
\subsection{Further Analysis}
\subsubsection{Tail Classes Accuracy} Our primary objective is incremental learning on imbalanced data, making it essential to focus on the results of tail classes. We conducted experiments on ImageNet-R with $\rho = 0.01$ across $10$ tasks, specifically evaluating the accuracy of tail classes at each incremental stage. The results are shown in Fig.\ref{fig4}. As the number of tasks increases, catastrophic forgetting is exacerbated as the imbalanced data distribution.  However, our method consistently maintains high performance, achieving $7.2 \%$ improvement over the previous SOTA after 10 tasks, demonstrating that our method effectively mitigates the challenges of LT-CIL. 
\begin{figure}[h]
\centering
\includegraphics[width=1\columnwidth]{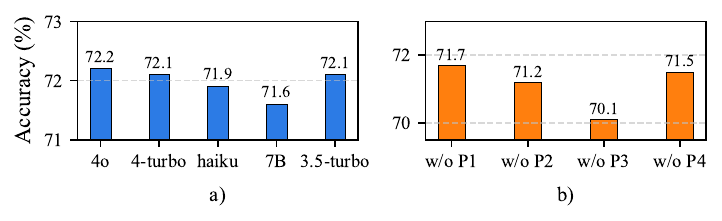} 
\caption{a) shows robustness of different LLMs and b) shows robustness of different prompt templates. All experiments were conducted on ImageNet-R with $\rho=0.01$ after 10 tasks.}
\label{fig5}
\end{figure}
\subsubsection{Robustness of Different LLMs} Our method relies on an LLM to generate task-specific text. To verify its robustness, we evaluate several different LLMs, including GPT-4o (4o), GPT-4-turbo (4-turbo), Claude-3.5-Haiku (haiku), all of which are commercial models, as well as Chat-Base-7B (7B), which is open-source and free. We conducted experiments on Imagenet-R, $\rho=0.01$, and $10$ tasks, and the experimental results are shown in Fig.\ref{fig5} a). It is evident that our method is robust to different LLMs; even when economic constraints are considered, an open-source free model can still generate sufficiently effective text to support our method.
\subsubsection{Robustness of Different Prompt Templates} To construct the SL-Tree, we designed four different prompt templates to guide the LLM in generating the desired text. To evaluate robustness with respect to prompt design, we conducted experiments on ImageNet-R with $\rho = 0.01$ and $10$ tasks, removing the text generated by the $i^{th}$ prompt while keeping all other conditions unchanged. The results are shown in Fig.\ref{fig5} b). The performance is affected when the corresponding text is deleted. When the coarse-grained text generated by prompt 1 has been removed, the accuracy drops to $71.7\%$, a decreased of $0.6\%$. Similarly, when the fine-grained descriptions from Prompt 2 or Prompt 3 are removed, the accuracy decreases to $71.2\%$ and $70.\%$, respectively. After removing Prompt 4, the final accuracy decreased to $71.5\%$, since prompt 4 cannot be applied to all classes and contributes selectively.
\begin{table}[h]
\centering
\caption{Experiments on ImageNet-LT.}
\label{tab5}
\begin{tabular}{lcccc}
\toprule
\multirow{2}{*}{Tasks} & \multicolumn{4}{c}{Method} \\ 
\cmidrule(lr){2-5}
& Upper Bound &CLIP Zero-Shot &Adapter Fine-tuning & Ours\\ 
\midrule
20 & 75.6 & 63.1 & 19.6 & 68.2 \\ 
50 & 75.6 & 63.1 & 17.4 & 66.7 \\ 
\bottomrule
\end{tabular}
\end{table}
\subsubsection{Experiments on Large Scale Dataset}
We conducted experiments on a large-scal dataset, ImageNet-LT \cite{liu2019large}, which was obtained by resampling from ImageNet-1K \cite{russakovsky2015imagenet} and contains 1000 different classes. We evenly divided ImageNet-LT into 20 and 50 independent tasks, requiring each task to learn 50 and 20 new classes, respectively. Classes with fewer than 400 samples were considered tail classes. Since most CIL and LT-CIL methods have not been evaluated on ImageNet-LT, we tested only the CLIP-based zero-shot, finetuning adapter, upper bound (joint learning) and our proposed method. The experimental results on VIT-B/16 are presented in the TABLE \ref{tab5}. It can be observed that our method remains effective on large-scale datasets and is not constrained by dataset size, demonstrating strong scalability and robustness in the long-tail class incremental learning setting.
\begin{figure}[h]
\centering
\includegraphics[width=0.95\columnwidth]{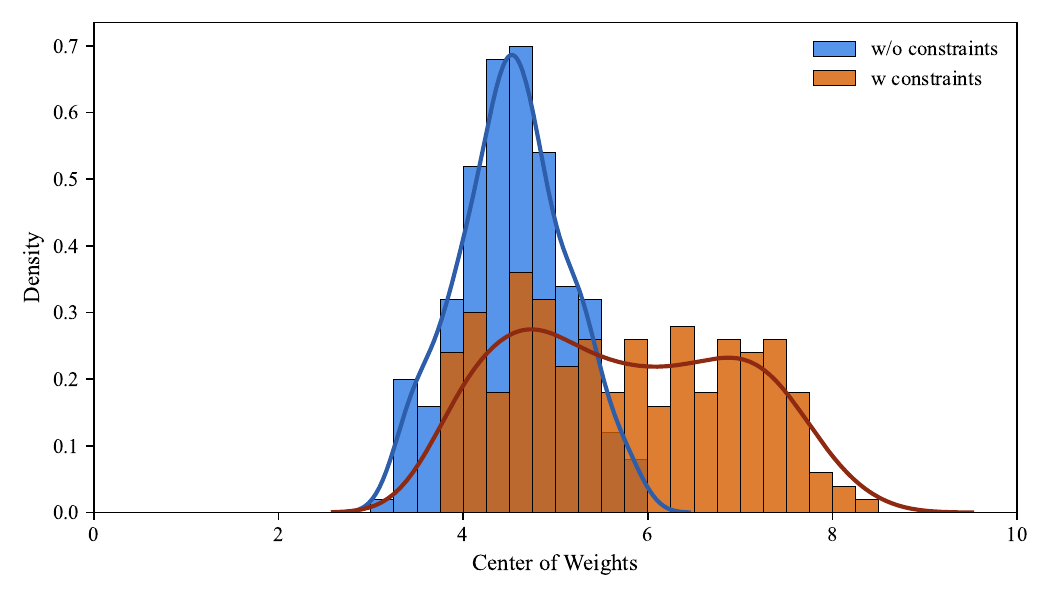} 
\caption{Density map of weights center. Experiments conducted on Imagenet-R, $\rho=0.01$, and $10$ tasks.}
\label{fig6}
\end{figure}
\subsubsection{Statistics of Weight Centers}
In the stratified adaptive language guidance, we introduced two additional constraints to regularize weight training, as formulated in Equations (7) and (9). Beyond the ablation studies presented in TABLE \ref{tab4}, we further visualized the learned weights to validate the effectiveness of these constraints. Specifically, experiments were conducted on ImageNet-R, 10 tasks with $\rho=0.01$, comparing models trained without the proposed constraints and models trained with them. After training, we computed the weight center of each class and plotted the centers of all 200 classes as a density map, as shown in Fig. \ref{fig6}. The results demonstrate that, after training, the weight centers are no longer concentrated in the middle layers but are instead adaptively redistributed across layers according to the class distribution. Moreover, the centers exhibit an overall tendency to shift toward higher layers. This observation aligns with the motivation behind the design of our constraints, which aim to prevent the weights from collapsing into a limited subset of layers and encourage tail classes to rely more heavily on higher-level semantic layers, thereby enhancing discriminability under long-tail incremental learning.
\subsubsection{Layer-wise Prediction Improvements} We calculate the difference in prediction probabilities for the correct label between two adjacent layers. The experiments were conducted on ImageNet-R with $\rho = 0.01$ and $10$ tasks. We present results on some tail classes in a bar chart format in Fig.\ref{fig7}, where the vertical axis represents the accuracy improvement and the horizontal axis corresponds to the layers being compared. Here, index $i$ means layer $i$ compares to layer $i+1$. As shown in the figure, because the $G^0$ of the SL-Tree provides a coarse-grained description of the entire task, performance improves when transitioning to $G^1$, which is generated by a fixed template. 
\begin{figure}[h]
\centering
\includegraphics[width=0.92\columnwidth]{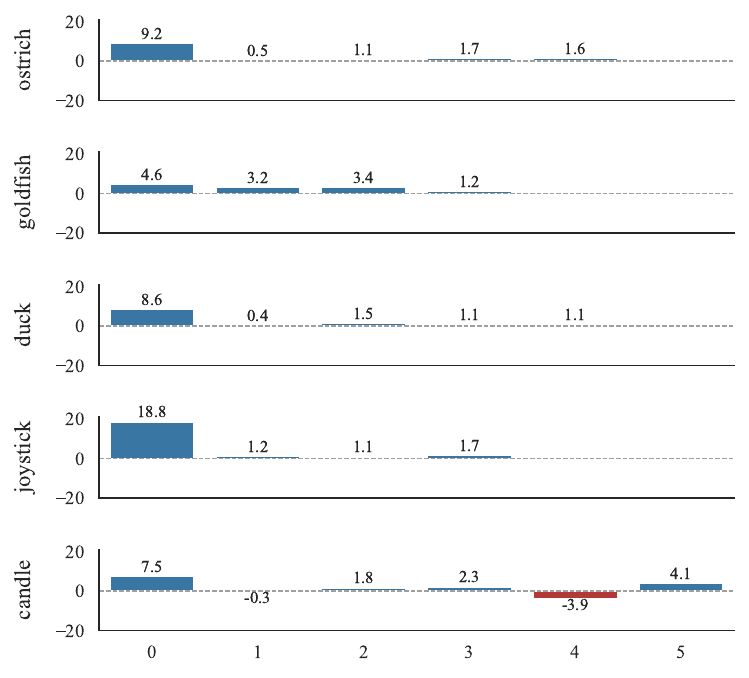} 
\caption{Performance difference between two adjacent layers. Experiments conducted on Imagenet-R, $\rho=0.01$, and $10$ tasks.}
\label{fig7}
\end{figure}
Furthermore, as the number of layers increases, the accuracy of most classes is positively correlated with depth, demonstrating why the SL-Tree—progressing from coarse-grained to fine-grained descriptions—outperforms fixed-template text features. However, there remain cases, such as candles, where the performance at a given layer is lower than that of the previous layer. This occurs because, as the depth of the SL-Tree increases, the generated text may randomly include content that is difficult for CLIP to interpret, such as “candle, which is characterized by various colors and scents.” This observation motivated the design of adaptive weights, rather than directly averaging across the SL-Tree.
\subsubsection{Effects on Tail Classes}
To verify the effect of stratified adaptive language guidance on tail classes, in addition to the overall tail classes' accuracy shown in Fig.\ref{fig4}, we further measured the per-class performance difference with and without stratified adaptive language guidance. The experiments were conducted on ImageNet-R, 10 tasks with $\rho=0.01$. The results, presented using box plots in Fig.\ref{fig8}, distinguish between head classes and tail classes. We observe that tail classes with fewer than 100 samples achieve substantially larger gains (mean $\Delta A = +16.6\%$) compared to head classes (mean $\Delta A = +4.1\%$), where $\Delta A$ denotes the accuracy improvement. This indicates that our method specifically enhances classification for data-scarce tail classes, cconsistent with its motivation to mitigate class imbalance in LT-CIL.
\begin{figure}[h]
\centering
\includegraphics[width=0.92\columnwidth]{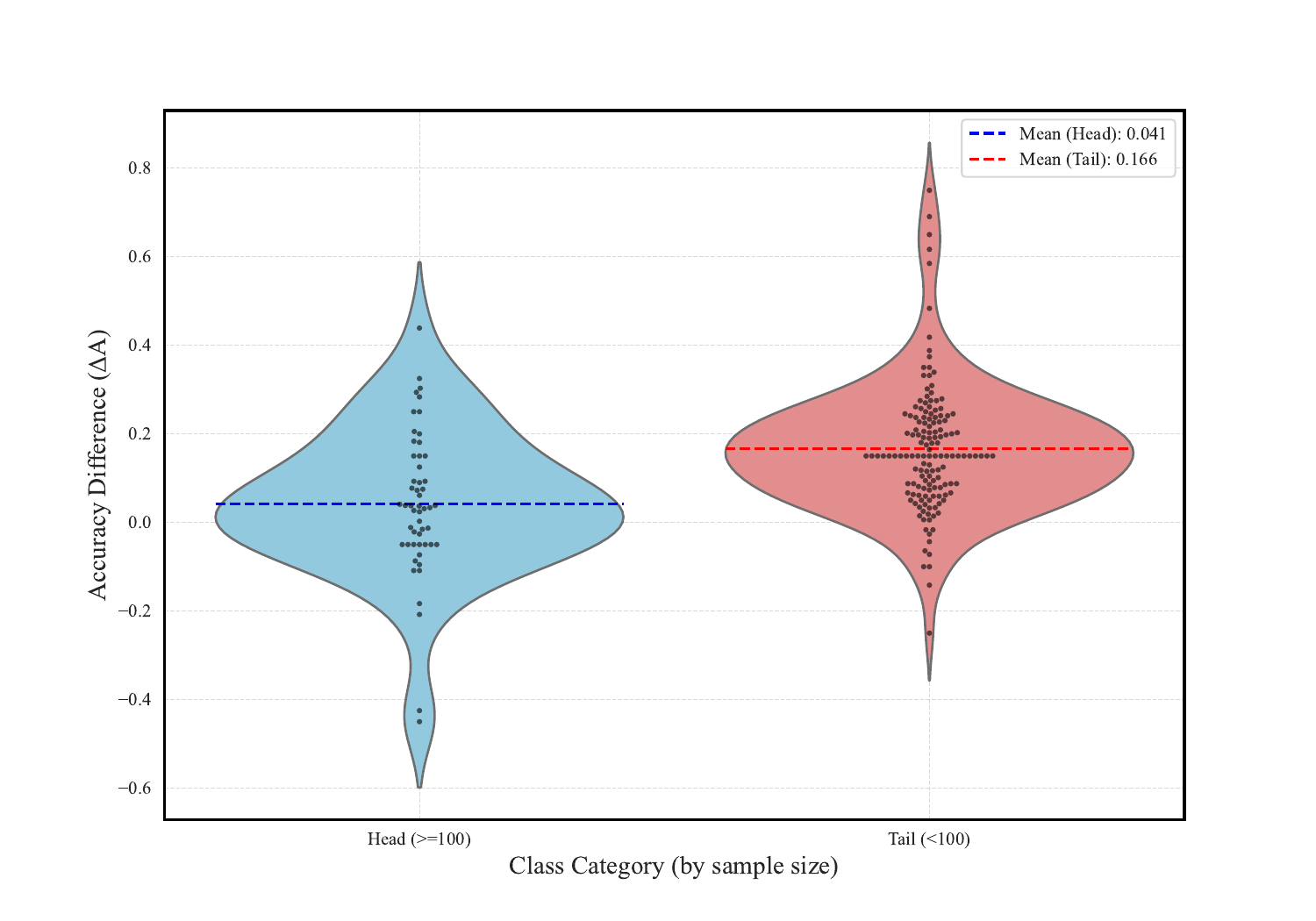} 
\caption{Effects on Tail Classes. Experiments conducted on Imagenet-R, $\rho=0.01$, and $10$ tasks.}
\label{fig8}
\end{figure}
\begin{table}[h]
\centering
\caption{Results on different pre-trained models. Experiments conducted on Imagenet-R, $\rho=0.01$, and $10$ tasks.}
\label{tab7}
\begin{adjustbox}{width=0.36\textwidth}
\begin{tabular}{lccc}
\toprule
\multirow{2}{*}{Method} & \multicolumn{3}{c}{Base model} \\ 
\cmidrule(lr){2-4}
& ViT-B/16 & ViT-L/14 & ViT-G/14 \\ 
\midrule
DAP & 64.8 & 74.8 & 77.9 \\ 
APART & 65.2 & 75.9 & 78.3 \\ \midrule
\rowcolor{gray!20}
\textbf{Ours} & \textbf{72.1} & \textbf{85.0} & \textbf{86.1} \\ 
\bottomrule
\end{tabular}
\end{adjustbox}
\end{table}
\subsubsection{Experiments on Different Pre-trained Model}
We conducted experiments on ImageNet-R with $\rho$ = 0.01 across 10 tasks using CLIP models of different sizes, VIT-B/16, VIT-L/14, and VIT-G/14. We evaluated different model structures against DAP and APART, two of the latest methods for LT-CIL. The experimental results are presented in Table \ref{tab7}. Our method consistently achieves the best performance across different CLIP model sizes, demonstrating strong scalability and robustness with respect to model capacity.
\begin{table}[h]
\centering
\caption{Long-Tail Experiments on ImageNet-R ($\rho=0.01$). \textbf{Bold} indicates the best result.}
\label{tab8}
\begin{adjustbox}{width=0.36\textwidth}
\begin{tabular}{lcc}
\toprule
\textbf{Method} & \textbf{All } & \textbf{Tail } \\ \midrule
Baseline (CLIP)                & 68.2 & 66.7 \\
Baseline$^\star$ (Adapter)     & 65.2 & 62.4 \\
\quad + LDAM                   & 68.5 & 66.7 \\
\quad + BalPoE                 & 68.5 & 66.3 \\
LFM+MMS                        & 70.1 & 69.5 \\
Baseline + SL-Tree             & 71.4 & 69.4 \\
Baseline$^\star$ + LTGC        & 73.0 & 70.2 \\ \midrule
\rowcolor{gray!20}
\textbf{Baseline$^\star$ + SL-Tree (Ours)} & \textbf{73.5} & \textbf{71.5} \\
\bottomrule
\end{tabular}
\end{adjustbox}
\end{table}
\subsubsection{Long-tail Experiments}
We conducted experiments on a long-tail (LT) dataset to evaluate the effectiveness of our method in addressing data imbalance. The experimental setup involved training and testing on the full ImageNet-R dataset with $\rho = 0.01$. We selected LFM+MMS \cite{liu2022long}, LDAM \cite{cao2019learning}, BalPoE \cite{aimar2023balanced}, and LTGC \cite{zhao2024ltgc} for comparison. The results are reported in Table \ref{tab8}. Baseline denotes direct CLIP zero-shot, while Baseline$\star$ refers to training the adapter; LDAM and BalPoE are both implemented with Baseline$\star$. We evaluate performance on all classes (All) and tail classes (Tail). The results show that fine-tuning the adapter is highly sensitive to data imbalance, leading to performance degradation and failing to achieve the expected improvement after transfer. By contrast, incorporating constraints from classic LT methods can mitigate the negative effects of imbalanced class distributions. Finally, SL-Tree leverages rich semantic information to compensate for data deficiencies and reduce the impact of task imbalances caused by long-tail data during model training.
\begin{table}[h]
\centering
\caption{Conventional class incremental learning experiments on CIFAR100, 10 tasks.}
\label{tab9}
\begin{adjustbox}{width=0.37\textwidth}
\begin{tabular}{l c}
\toprule
\textbf{Method} & \textbf{All Accuracy (\%)} \\ 
\midrule
CODAPrompt & 76.7 \\
GMM & 78.0 \\
RAPF & 78.5 \\
MG-CLIP & 79.4 \\
\midrule
\rowcolor{gray!20}
Ours (SL-Tree, $\rho=0.1$) & \textbf{81.2} \\
\rowcolor{gray!20}
Ours (SL-Tree, $\rho=0.01$) & 81.0 \\
\rowcolor{gray!20}
Ours (SL-Tree, $\rho=1$) & 80.4 \\
\bottomrule
\end{tabular}
\end{adjustbox}
\end{table}
\subsubsection{Conventional Class Incremental Learning Experiments}
Since most of the comparison methods in TABLE \ref{tab1}-\ref{tab3} are originally designed for conventional class incremental learning (CIL), we further conducted experiments under the standard CIL setting. For our method, we considered two scenarios: 1) using the SL-Tree generated under imbalanced conditions (with $\rho=0.1$ and $\rho=0.01$), while keeping the rest of the method unchanged; and 2) removing prompt template 4 and adopting the remaining templates to uniformly guide the LLM in generating text descriptions (corresponding to SL-Tree with $\rho=1$). The results on  CIFAR100  with 10 tasks are reported in Table \ref{tab9}. It can be seen that when the data is balanced, better performance can be obtained by directly using our previously generated SL-Tree. However, due to the inconsistency between the balanced training distribution and the imbalanced semantic structure used to generate the SL-Tree, the improvement is still smaller than in the imbalanced case. And when we migrate the strategy of generating text easily to balanced data, it only gives slight improvement compared to the previous SOTA method due to the missing cycling generation. Therefore, adapting our method more effectively to the balanced data distribution remains a worthwhile research direction.
\begin{figure}[h]
    \centering
    \subfloat[]{
        \includegraphics[width=0.23\textwidth]{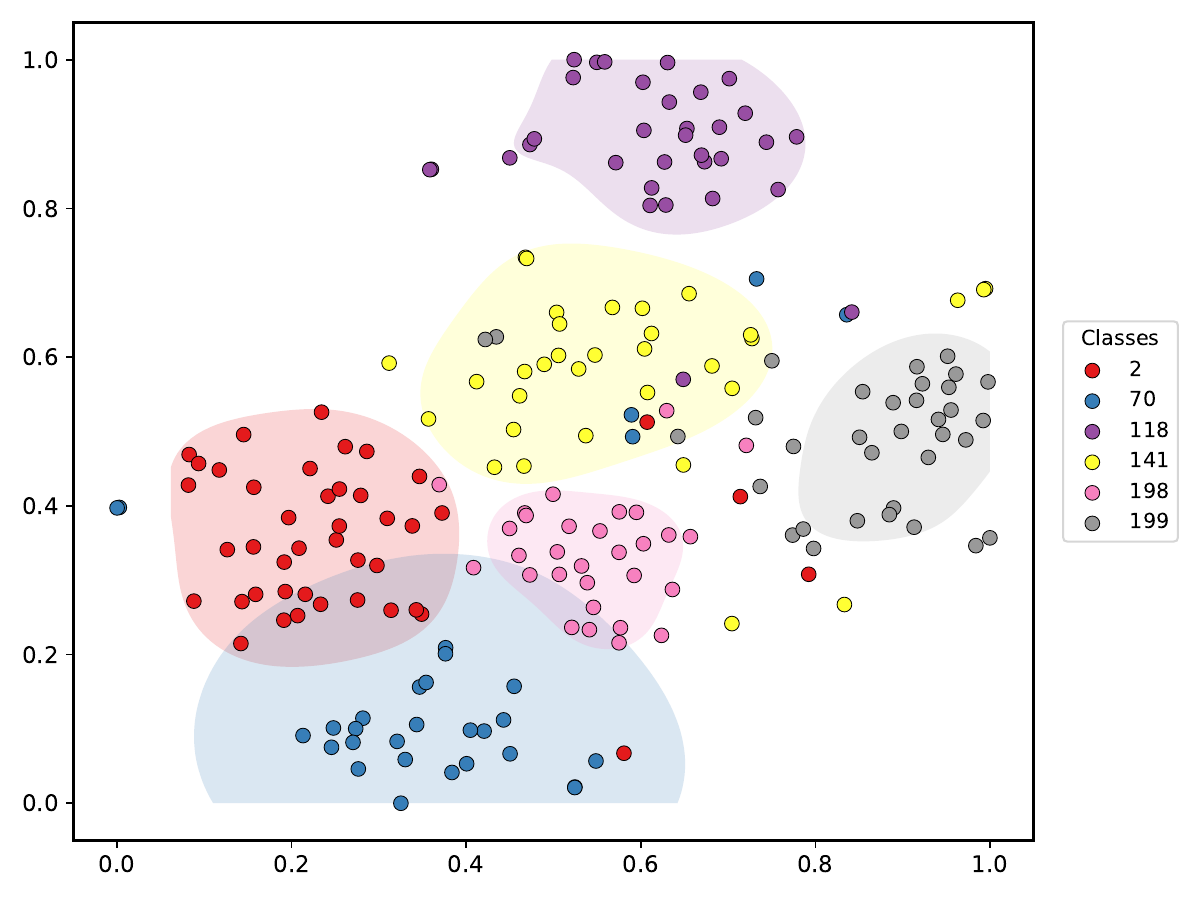}
    }
    \subfloat[]{
        \includegraphics[width=0.23\textwidth]{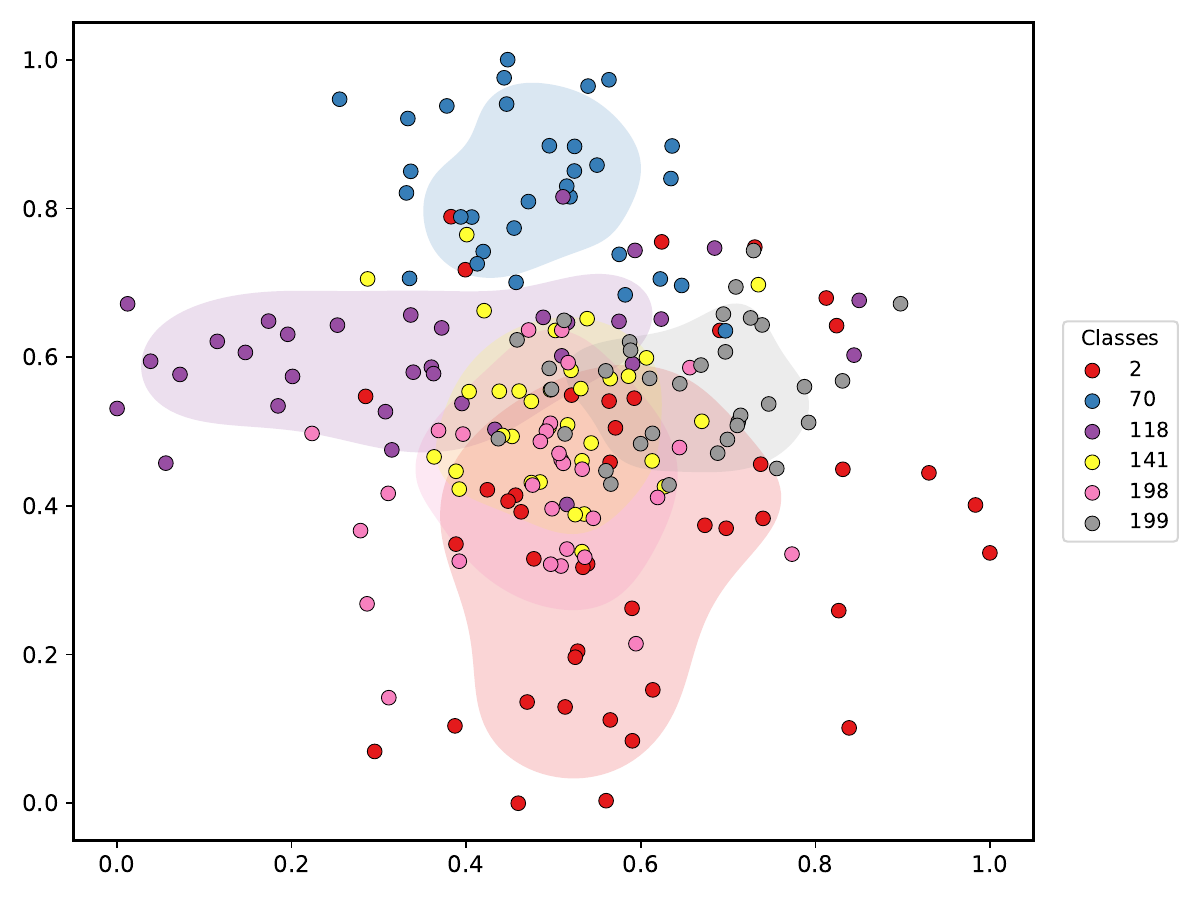}
    }

    \subfloat[]{
        \includegraphics[width=0.23\textwidth]{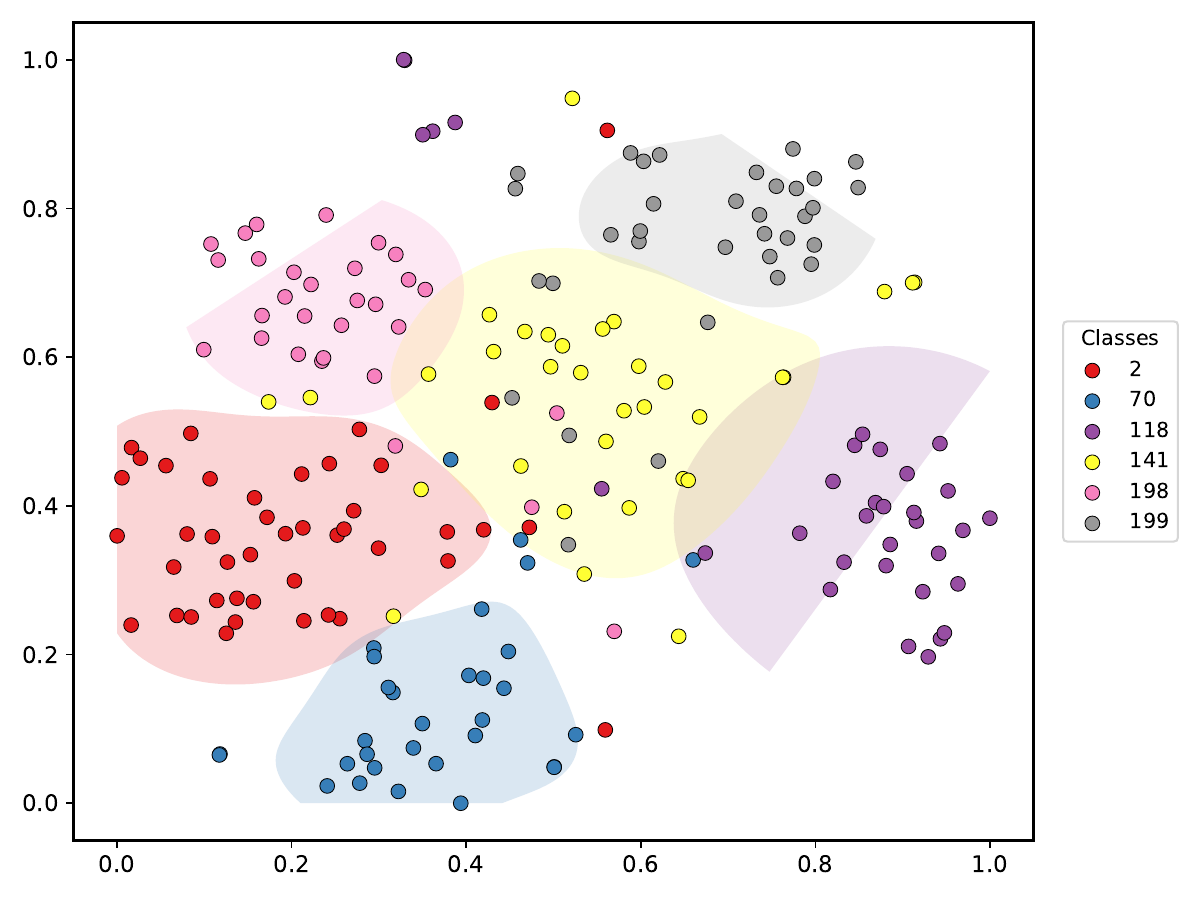}
    }
    \subfloat[]{
        \includegraphics[width=0.23\textwidth]{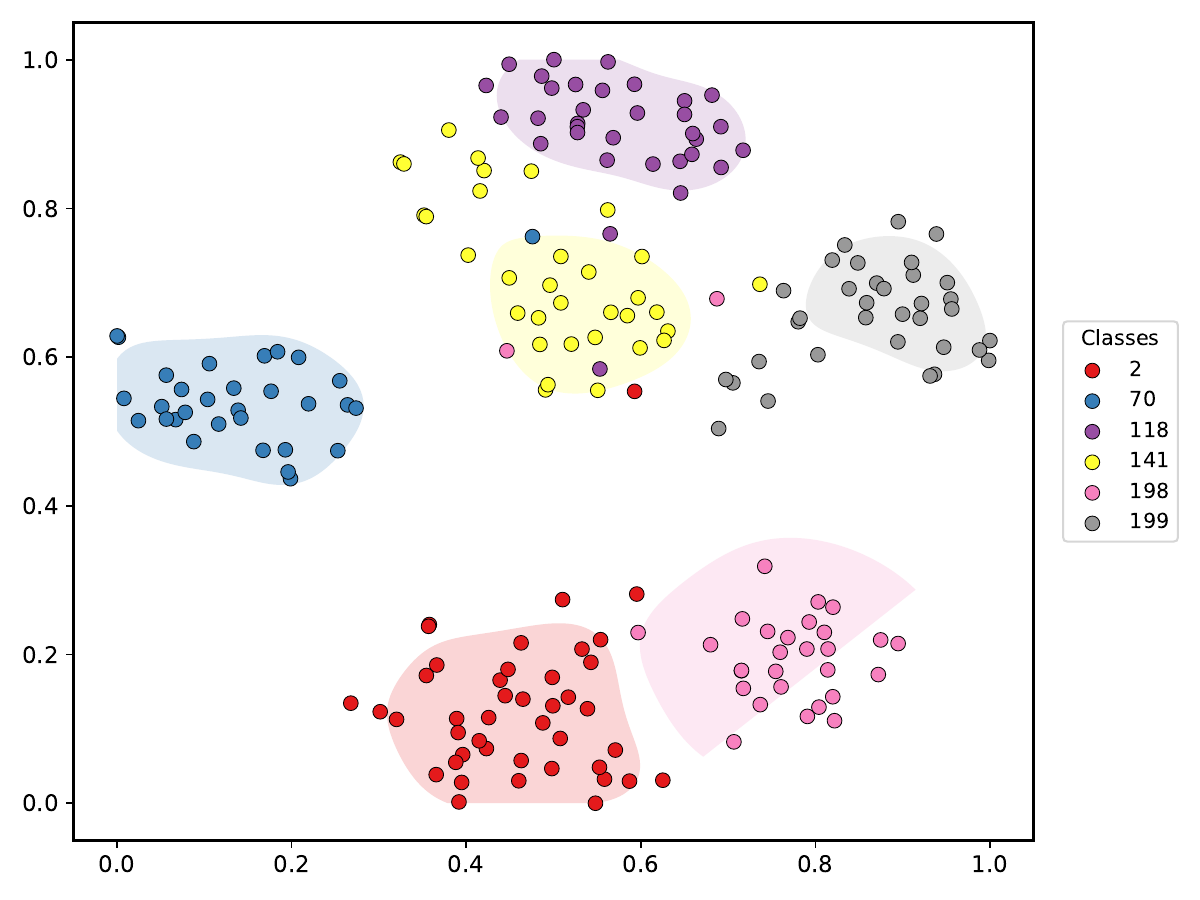}
    }
    \caption{t-SNE visualization of different classes. Experiments conducted on ImageNet-R, $\rho=0.01$, and $10$ tasks. a) zero-shot; b) fine-tuned adapter; c) fine-tuned adapter with knowledge distillation (KD); d) fine-tuned adapter with KD and the proposed stratified alignment language guidance.}
    \label{fig9}
\end{figure}
\subsubsection{t-SNE Visualization of Different Classes}
To further validate our proposed method, we conducted experiments using stratified alignment language guidance under four configurations: (1) zero-shot; (2) fine-tuned adapter; (3) fine-tuned adapter with knowledge distillation (KD); and (4) fine-tuned adapter with KD and the proposed adaptive language guidance. The experiments were conducted on ImageNet-R, $\rho=0.01$ with 10 tasks. We selected head classes 2, 118, and 198, and tail classes 20, 141, and 199 from the test set. The t-SNE visualization results are shown in Fig.\ref{fig9}. As observed, direct fine-tuning leads to a pronounced forgetting phenomenon on tail classes, resulting in highly entangled and poorly separated class distributions in the feature space. Incorporating knowledge distillation slightly mitigates this issue; however, compared with the zero-shot, the classification performance on tail classes remains suboptimal. When our proposed stratified alignment language guidance is introduced, the classification performance—particularly for tail classes—improves substantially, leading to clearer class boundaries and more compact feature clusters.
\begin{figure}[t]
    \centering
    \subfloat[]{
        \includegraphics[width=0.22\textwidth]{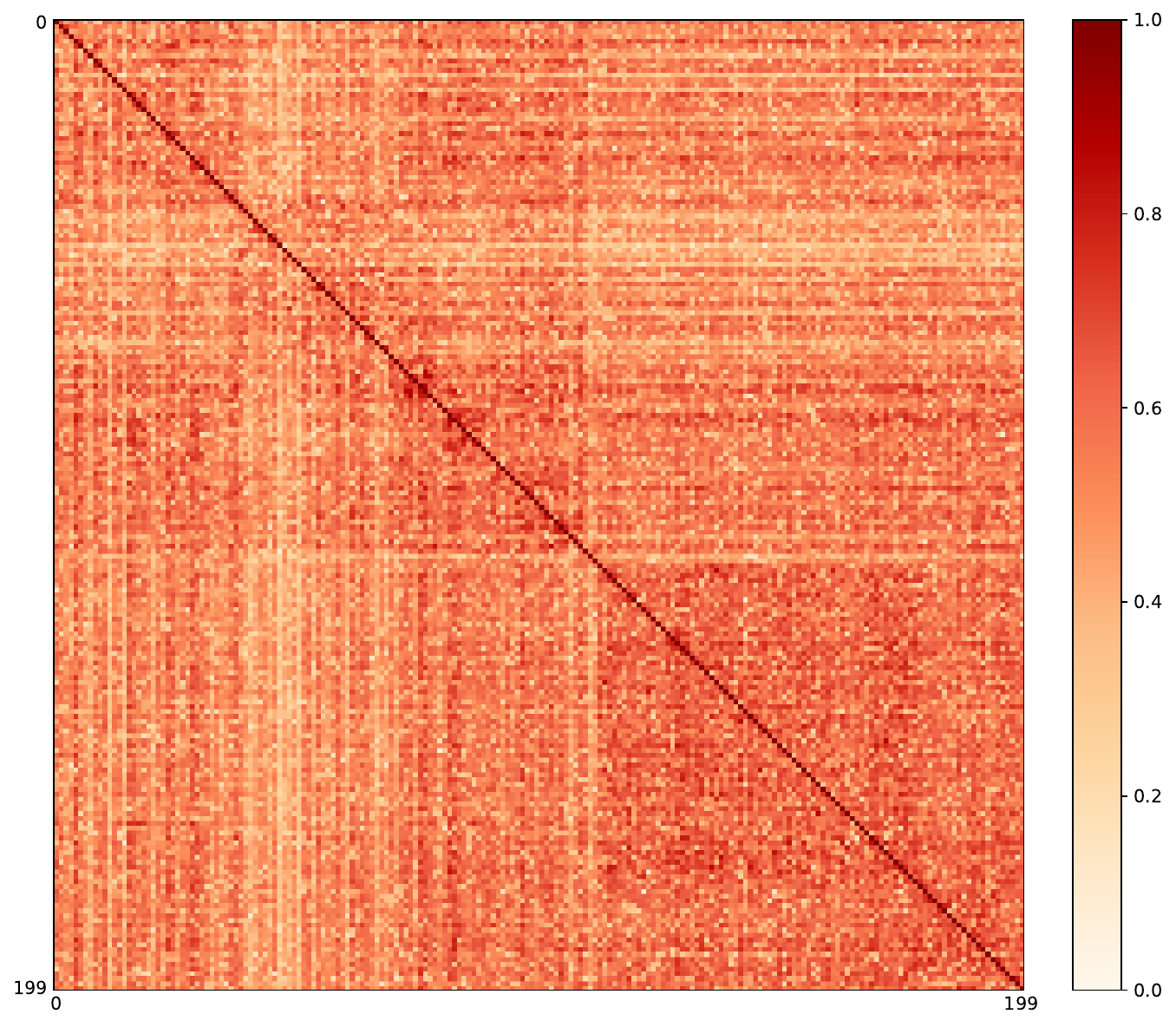}
    
    }
    \subfloat[]{
        \includegraphics[width=0.22\textwidth]{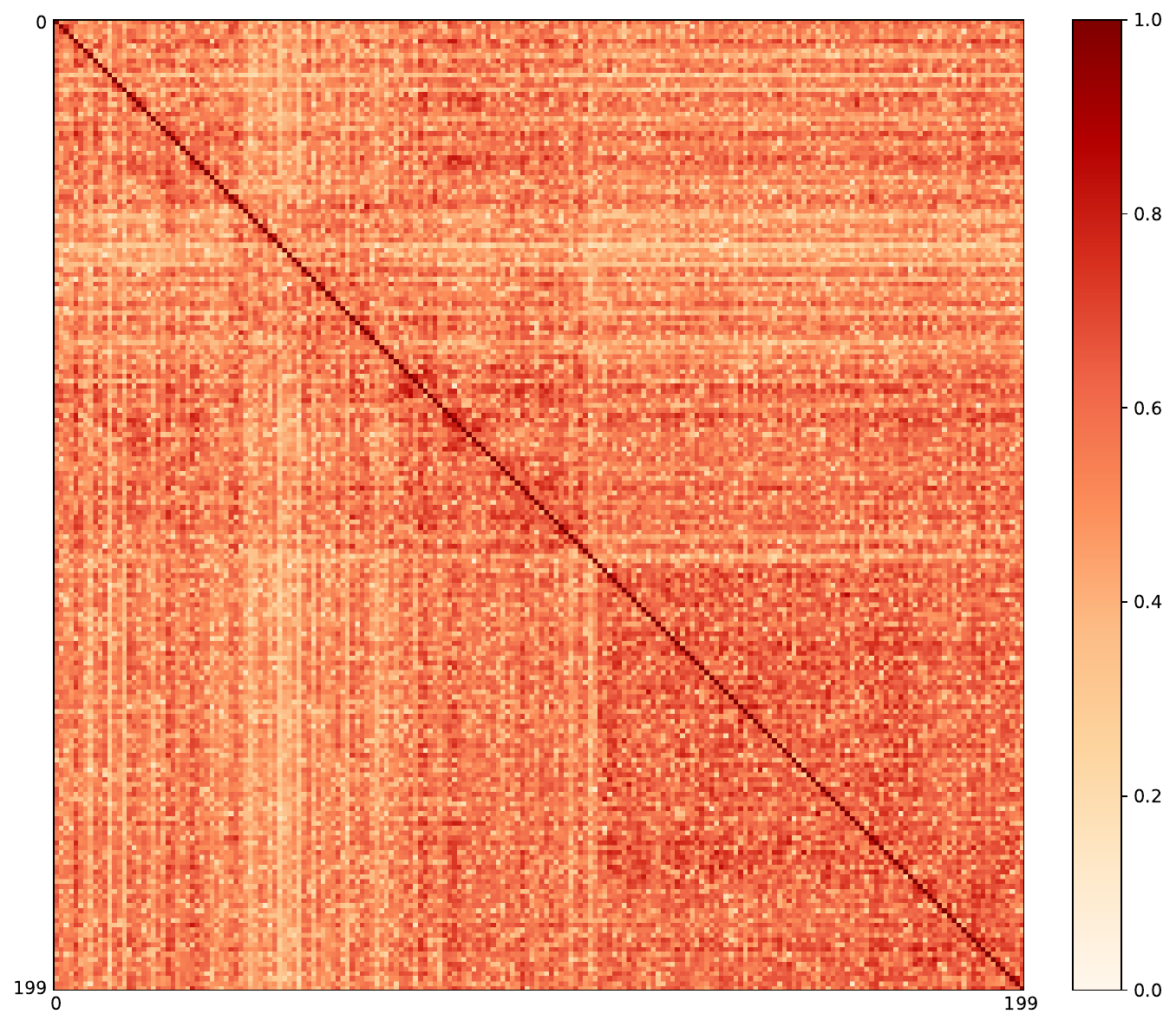}
      
    }


    \subfloat[]{
        \includegraphics[width=0.22\textwidth]{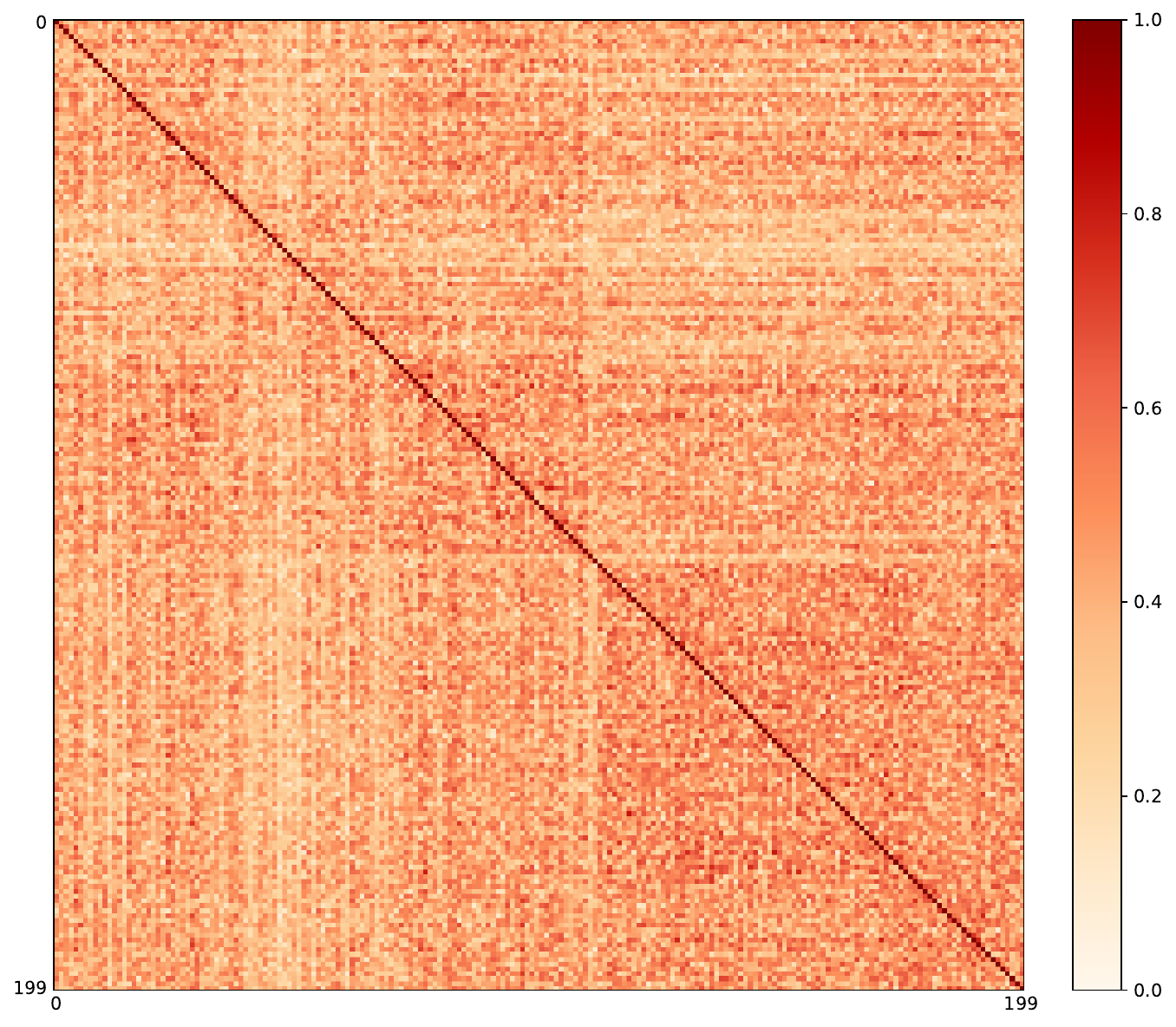}
        
    }
    \subfloat[]{
        \includegraphics[width=0.22\textwidth]{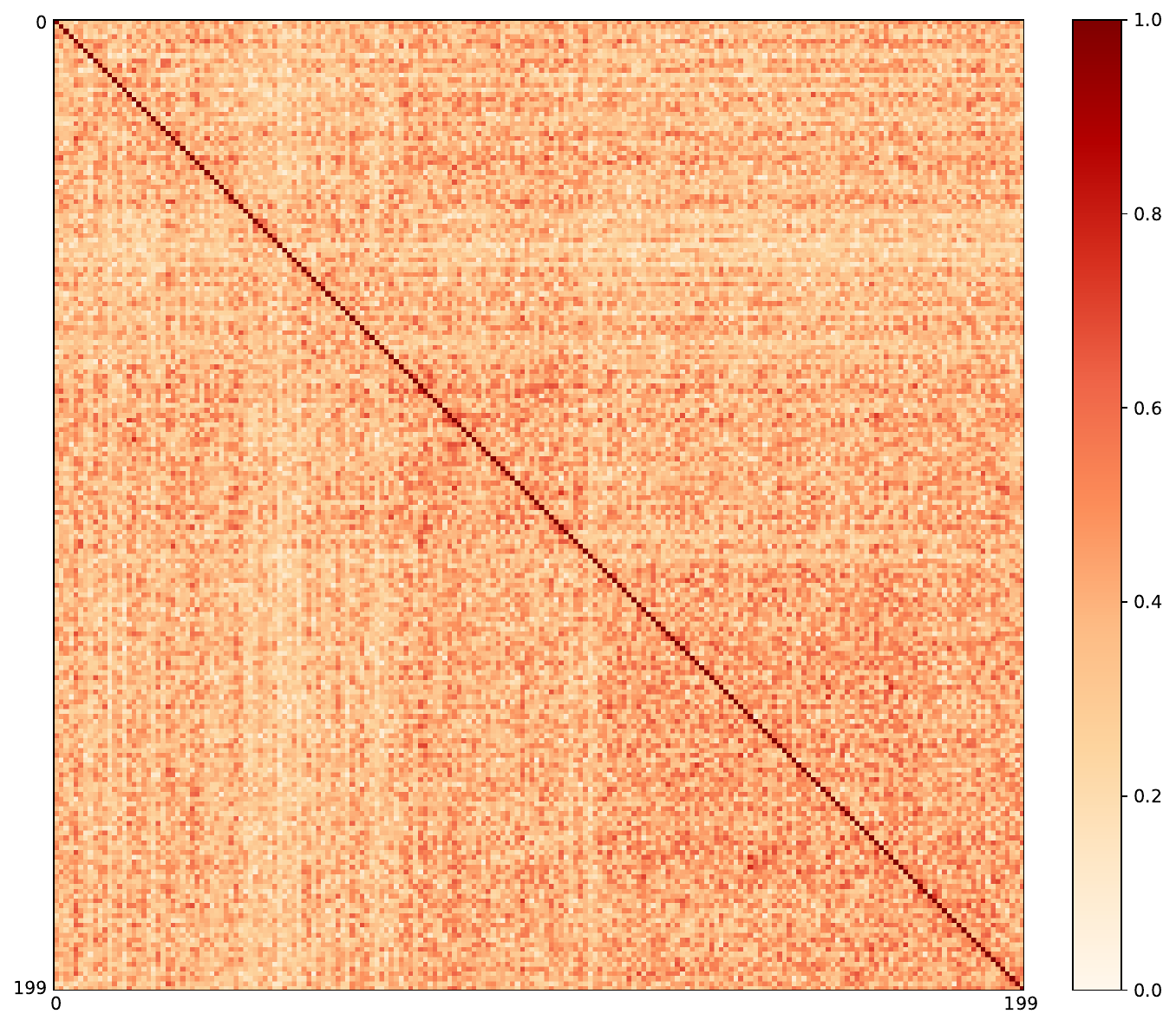}
   
    }
    \caption{Similarity heatmap of text featuress. Experiments conducted on ImageNet-R, $\rho=0.01$, and $10$ tasks, a) is the fixed text; b) duplicating the fixed text; c) is the mean of text from different layers of SL-Tree; d) is the linearly weighted text from different layers of SL-Tree.}
    \label{fig10}
\end{figure}
\subsubsection{Similarity Heatmap of Text Features}
To further validate our proposed stratified adaptive language guidance, we conducted similarity experiments on text features. Specifically, the experiments were conducted on ImageNet-R with 10 tasks under $\rho=0.01$. Four methods of using text were evaluated: 1) texts generated with a fixed template `a photo of'; 2) duplicating the texts in 1) to match the number of texts in SL-Tree and averaging the corresponding text features; 3) averaging the text features extracted from different layers of the SL-Tree; and 4) linearly weighting the trained parameters associated with text features from different SL-Tree layers. We visualize cosine similarities between text features of different classes in a heatmap. As shown in Fig.\ref{fig10}. The similarities among text features generated by the proposed SL-Tree are substantially lower than those produced using the fixed template, indicating a richer and more discriminative semantic representation. In addition, employing trainable weights yields better performance than directly averaging features from multiple layers. Furthermore, simply expanding the number of texts to the same scale as SL-Tree does not improve performance, highlighting that quantity alone cannot replace semantic diversity.
\begin{figure}[t]
\centering
\includegraphics[width=0.96\columnwidth]{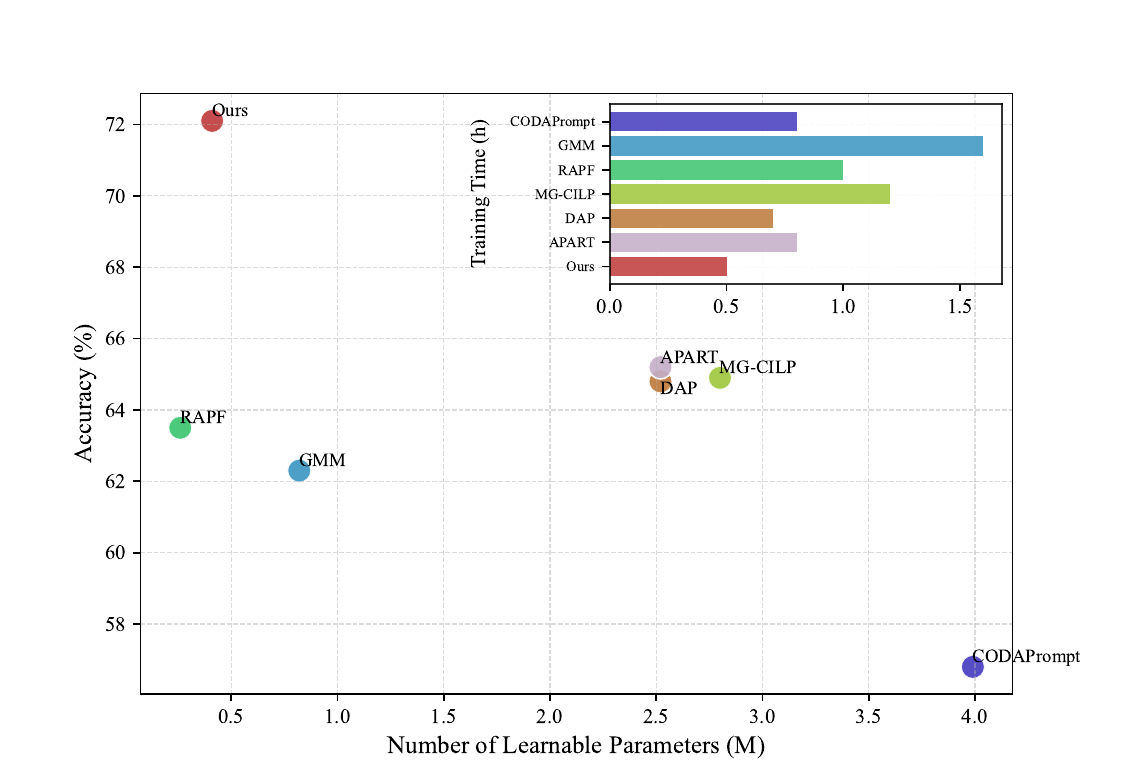} 
\caption{Experimental consumption. Experiments conducted on Imagenet-R, $\rho=0.01$, and $10$ tasks.}
\label{fig11}
\end{figure}
\begin{figure}[h]
    \centering
    \subfloat[$\lambda_1$.]{
        \includegraphics[width=0.23\textwidth]{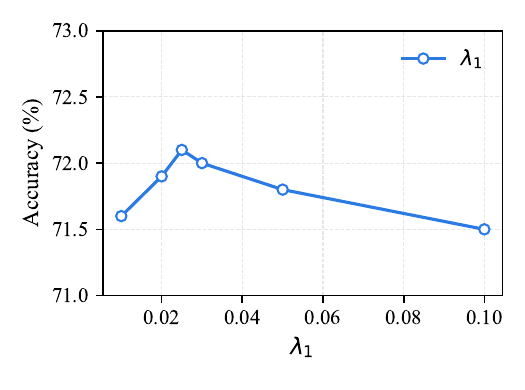}
      
    }
    \subfloat[$\lambda_2$.]{
        \includegraphics[width=0.23\textwidth]{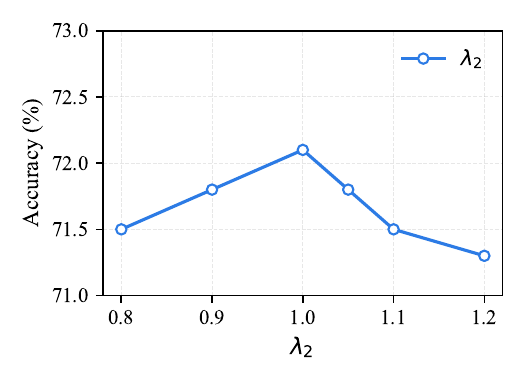}
  
    }


    \subfloat[$\lambda_3$.]{
        \includegraphics[width=0.23\textwidth]{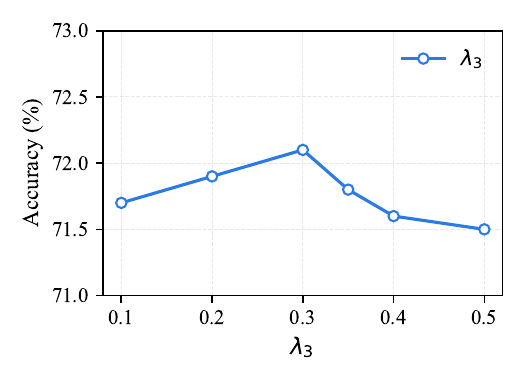}
      
    }
    \subfloat[$\lambda_4$.]{
        \includegraphics[width=0.23\textwidth]{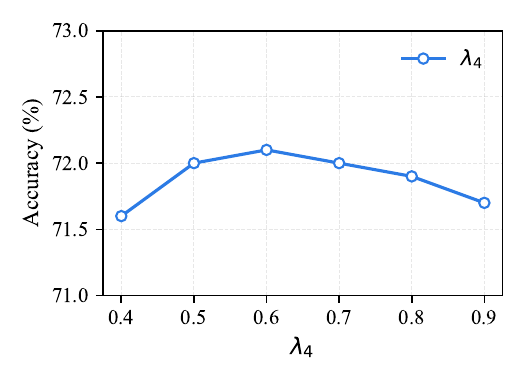}
    }
    \caption{Sensitivity analysis of hyperparameters. Experiments conducted on ImageNet-R, $\rho=0.01$, and $10$ tasks.}
    \label{fig12}
\end{figure}
\subsubsection{Experimental Consumption}
We analyzed the number of trained parameters and training time required by different methods on ImageNet-R with $\rho = 0.01$ across 10 tasks to compare their computational cost. The results are presented in Fig.\ref{fig11}. In the bubble chart, the horizontal axis represents the number of parameters trained by each method, while the vertical axis represents accuracy. The accompanying histogram illustrates the training time of each method. As shown, our method achieves the highest accuracy while requiring a small number of parameters (0.41M) and the least training time (0.5 hours). This demonstrates that our method is not only effective but also highly efficient, offering strong performance with minimal computational overhead.
\subsubsection{Sensitivity Analysis of Hyperparameters}
In Equation (18), both losses and constraints are multiplied by different weights. We provide a comprehensive analysis of the hyperparameters, and all experiments were conducted on ImageNet-R with an imbalance ratio $\rho=0.01$ and 10 tasks. The results of hyperparameters $\lambda_1$ - $\lambda_4$ are shown in Fig.\ref{fig12} a) - d), and the final value of $\lambda_1 = 0.025$, $\lambda_2 = 1$, $\lambda_3 =0.3$ and $\lambda_4 =0.6$. Our method is stable to different hyperparameters.
\section{Conclusion}
In this paper, we propose a stratified language tree and two parallel language guidances, stratified adaptive language guidance and stratified alignment language guidance. Specifically, we first guide a large language model to generate a stratified language tree containing multi-scale semantic information. Then, adaptive language guidance introduces an adaptive trainable weights  to leverages text descriptions to compensate for the lack of visual information in tail classes. Meanwhile, alignment language guidance exploits the stability of semantic information to constrain the model's optimization, thereby reducing the influence of insufficient visual data on previously learned classes and mitigating catastrophic forgetting. By jointly leveraging these two complementary guidance mechanisms, our method effectively addresses the key challenges of long-tail class incremental learning (LT-CIL) and achieves superior performance across various benchmarks.

\bibliographystyle{IEEEtran}
\bibliography{tpami.bib}


 





\end{document}